\documentclass[runningheads]{llncs}
\usepackage[T1]{fontenc}
\usepackage{amsfonts,amssymb,amsmath}
\usepackage{graphicx}
\usepackage{subfigure}
\usepackage{bbding}
\usepackage{graphicx}
\begin{document}

\title{SpineCLUE: Automatic Vertebrae Identification Using Contrastive Learning and Uncertainty Estimation}
\titlerunning{Automatic Vertebrae Identification Using SpineCLUE}

\author{Sheng Zhang\inst{1} \and Minheng Chen\inst{2} \and Junxian Wu\inst{1} \and Ziyue Zhang\inst{1} \and Tonglong Li\inst{1} \and Cheng Xue\inst{1,2,3(}\Envelope\inst{)} \and Youyong Kong\inst{1,2,3(}\Envelope\inst{)}}

\institute{Jiangsu Provincial Joint International Research Laboratory of Medical Information Processing,
Southeast University, Nanjing 210000, China \and
School of Computer Science and Engineering, Southeast University, Nanjing 210000, China \and Key Laboratory of New Generation Artificial Intelligence Technology and Its Interdisciplinary Applications (Southeast University), Ministry of Education, Nanjing 210000, China\\
\email{cxue@seu.edu.cn, kongyouyong@seu.edu.cn}}

\authorrunning{S.Zhang et al.}
%
\maketitle             
%
\begin{abstract}
Vertebrae identification in arbitrary fields-of-view plays a crucial role in diagnosing spine disease. Most spine CT contain only local regions, such as the neck, chest, and abdomen. Therefore, identification should not depend on specific vertebrae or a particular number of vertebrae being visible. Existing methods at the spine-level are unable to meet this challenge.
In this paper, we propose a three-stage method to address the challenges in 3D CT vertebrae identification at vertebrae-level. By sequentially performing the tasks of vertebrae localization, segmentation, and identification, the anatomical prior information of the vertebrae is effectively utilized throughout the process. 
Specifically, we introduce a dual-factor density clustering algorithm to acquire localization information for individual vertebra, thereby facilitating subsequent segmentation and identification processes. 
In addition, to tackle the issue of inter-class similarity and intra-class variability, we pre-train our identification network by using a supervised contrastive learning method. To further optimize the identification results, we estimated the uncertainty of the classification network and utilized the message fusion module to combine the uncertainty scores, while aggregating global information about the spine.
Our method achieves state-of-the-art results on the VerSe19 and VerSe20 challenge benchmarks. Additionally, our approach demonstrates outstanding generalization performance on an collected dataset containing a wide range of abnormal cases.

\end{abstract}

\section{Introduction}
\label{Sec. 1}
Automatic vertebrae identification from CT images with arbitrary fields-of-view (FOVs) is essential for spine disease diagnosis, surgical treatment planning, and postoperative evaluation. Doctors can analyze the shape and angle of individual vertebra based on the recognition results to facilitate early diagnosis and surgical treatment of degenerative diseases, vertebrae deformities, spinal tumors, compression fractures, and other spinal pathologies \cite{YAO201616,Forsberg_2013}.

\begin{figure}[t]
\centering
\includegraphics[width=0.95\textwidth]{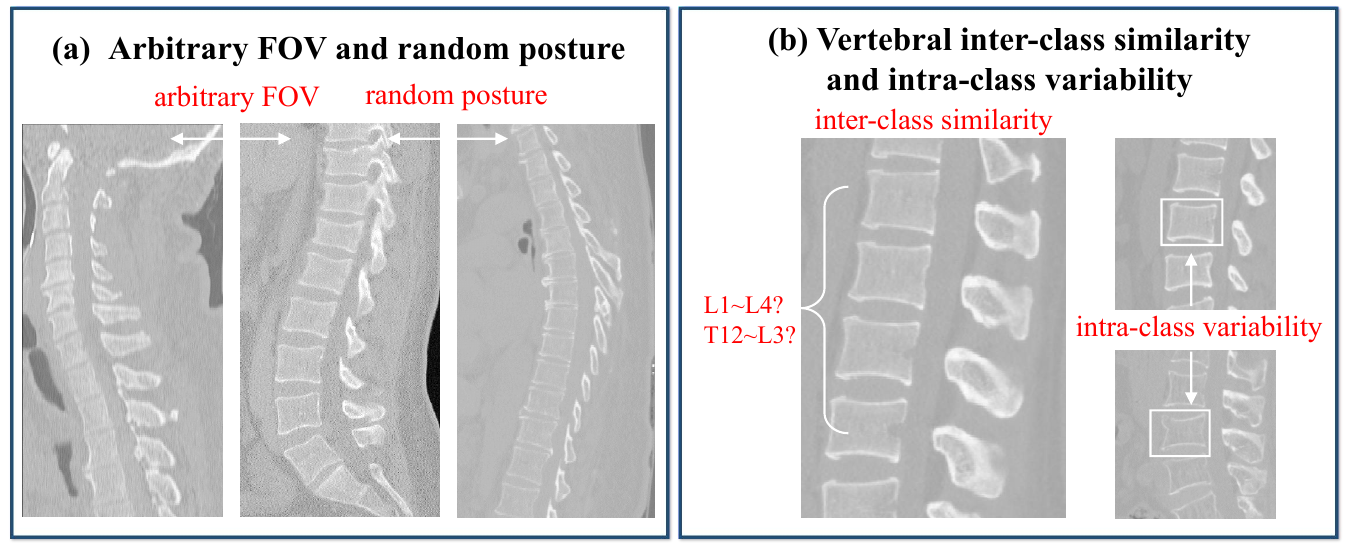}
\caption{Example illustration of challenges encountered in vertebrae identification. \textbf{a)} The first and second vertebrae have different FOVs(cervical and lumbar). And there are differences in vertebrae posture between the second and third images. \textbf{b)} The shapes of adjacent vertebrae frequently exhibit similarities. As depicted in the left picture, it is difficult to distinguish whether these vertebrae are L1-L4 or T12-L3. Meanwhile, the spines belonging to the same category (masked by white boxes in the right two figures) also have certain variations. These two characteristics makes the process of extracting features of a specific category more difficult.}
\label{fig:challenge}
\end{figure}

Lots of effort is trying to achieve highly accurate vertebrae identification, but the performances are not ideal due to the variable morphology and posture of the vertebrae.
As shown in Fig.~\ref{fig:challenge}, the primary challenges associated with the task stem from the following two difficulties: \textbf{1)} Arbitrary FOVs and random posture: in clinical practice, the FOV of CT varies with scanning equipment or scanning configurations, which results in large variations in the position and shape of the vertebrae in the captured images. At the same time, the random posture of the vertebrae can also lead to variation in the position and orientation of the vertebrae. \textbf{2)} Vertebral inter-class similarity and intra-class variability: vertebral inter-class similarity refers to the fact that the shapes and structures of a certain two upper and lower neighboring vertebrae within the same patient can exhibit significant similarities. This similarity is particularly evident in the lumbar and thoracic regions. Meanwhile, intra-class variability means that the same vertebra may display variations in shape and structure across different patients, which can be attributed to various spinal conditions and age-related factors. Both of them contribute to the heightened challenge associated with the vertebrae identification.

To address the first challenge, method for vertebrae identification necessitates the ability to effectively handle diverse image FOVs. This implies that no prior assumptions should be made about the quantities and classifications of vertebrae. Therefore, the task can be conceptualized as a specialized instance segmentation task, wherein the vertebrae targets exhibit continuity and high similarity. However, the end-to-end instance segmentation method cannot effectively integrate the sequential information of the spine into the recognition process, and it is difficult to achieve satisfactory performance solely based on the local image information of the spine. So many methods have emerged to decompose the task into multiple stages that concentrate on integrating prior anatomical information (top and bottom consecutive, labels not duplicated, etc.) of the spine into the recognition process \cite{Liao2018,chen2020hmm,meng2023vertebrae}. However, most multi-stage methods execute tasks in isolation, without leveraging the information between modules. This results in the model missing many valuable global insights within each model, which are highly beneficial for identification. For instance, localization can provide spatial information for segmentation, preventing the segmentation of adjacent vertebrae within the bounding box. The segmentation results can then refine the identification bounding box, while using the segmentation mask as input to the identification model can provide more distinct boundaries and morphology. Our methodology utilizes a multi-stage framework, which breaks down the task into three sub-tasks: localization, segmentation and identification. What distinguishes our approach from others is the deliberate integration of modules, enabling each subsequent task to efficiently utilize information obtained from the preceding task. 

Vertebrae localization models trained using common datasets often fail to detect abnormal vertebrae such as scoliosis and kyphosis, which leads to incomplete subsequent segmentation and identification results. In our study, we transform the 3D localization problem into a 2D one, thereby reducing the localization difficulties associated with 3D anomalous poses. Additionally, it is unnecessary to predict the mass centers of the vertebrae. Instead, the vertebrae target boxes or likelihood centers are more beneficial in clinical practice. We utilize a fast detector to localize vertebrae on 2D slices of volumes to obtain the bounding boxes of vertebrae on each slice. However, the bounding boxes of these slices are partially noisy, and the size and position of the bounding boxes of the same vertebrae on different slices are various. Therefore, we propose a dual-factor density clustering method to cluster the bounding boxes between different slices without assumption of the number of categories, while eliminating the noise. The localization results can be used to crop the segmentation bounding boxes and provide a prior position information to the segmentation model, thereby enhancing its capability to effectively differentiate the target vertebrae from the potential presence of multiple vertebrae within the boxes.

Additionally, it is imperative to develop an identification strategy that effectively addresses the issue of inter-class similarity and intra-class variability. With the widespread exploration of representation learning, contrastive learning methods have garnered significant attention and interest among researchers \cite{Wu_2018_CVPR,pmlr-v119-chen20j,SupCon}. By employing the construction of positive pairs and negative pairs, contrastive learning trains a representation learning model, ensuring that positives are projected closely together in the embedding space, while negatives are projected far apart. This aligns well with our definition of inter-class similarity and intra-class variability, corresponding to hard negatives and hard positives respectively \cite{Shrivastava_2016_CVPR}. Specifically, we utilize supervised contrastive learning to generate positive and negative pairs based on vertebrae labels. Our method is the first to apply supervised contrastive learning pre-training and projection head fine-tuning for vertebrae identification and has achieved promising outcomes.

Although supervised contrastive learning pre-training method can provide significant assistance in identification, the prediction of the network is not deterministic, making the results subject to a high degree of uncertainty. Generally, uncertainty is divided into two types \cite{Gawlikowski2021}: \textbf{1)} Data (Aleatoric) uncertainty: it describes the inherent noise in the data, which represents unavoidable errors. \textbf{2)} Model (Epistemic) uncertainty: it refers to the potential inaccuracy of the model's predictions due to factors such as inadequate training or insufficient training dataset, and it is unrelated to specific data. For vertebrae identification, uncertainty refers mainly to the second one, which can lead to the fact that if a vertebra is incorrectly identified, its neighbouring vertebrae will be affected when fusing message using certain method. This type of uncertainty can be mitigated or resolved through targeted adjustments \cite{Devries2018,Yu2019,Yang_2021_CVPR}. Therefore, we aim to assess the accuracy of predictions by estimating the uncertainty scores of the network and utilize these scores to weight the global information integrated into the identification of individual vertebrae, thereby ensuring effective information is provided to the identification process.
In conclusion, we propose a vertebrae-level three-stage framework for \textbf{Spine} identification using \textbf{C}ontrastive \textbf{L}earning and \textbf{U}ncertainty \textbf{E}stimation called \textbf{SpineCLUE} to hierarchically achieve vertebrae identification in this paper. Our main contributions of this study can be summarized as follows:

\begin{itemize}
\item We propose a novel vertebrae-level vertebrae identification for FOV that performs localization, segmentation, and identification sequentially. In this approach, each task utilizes information provided by the preceding task to facilitate the identification process.

\item In order to improve the stability of localization and prevent the failure caused by abnormal vertebrae in the 3D space, we use a detector to locate the likelihood centers of the vertebrae and propose dual-factor density clustering to obtain the localization results. 

\item To address the suboptimal identification results arising from inter-class similarity and intra-class variability, we utilize supervised contrastive learning to increase the separation between different categories.

\item We design an uncertainty estimation module to integrate prior information along the spine sequence and weighted the fusion of messages between vertebrae using the uncertainty from the identification model, providing clue for vertebrae identification.
\end{itemize}

\section{Related work}
\subsection{Vertebrae identification}
Previous studies on vertebrae identification methods are based on active contours as well as graph theory based methods \cite{Hammermik2015,Athertya2016}. These methods rely too much on prior knowledge and perform poorly on abnormal spine images. With the widespread use of machine learning techniques, many studies have started to adopt a multi-stage approach to accomplish vertebrae identification, where each stage is accomplished using an end-to-end machine learning approach. Chu et al.\cite{chu2015} utilized a learning-based random forest to regress vertebrae centers and cropped regions of interest (ROIs) for the segmentation of individual vertebra based on the centers. Similarly, Wang et al.\cite{wang2019} used an auto-encoder to extract contextual features of the spine and then implemented vertebrae identification through structured random forests.

Recently, deep learning methods have gained popularity, many researchers are attempting to integrate anatomical priors for vertebrae identification into the deep learning process. For example, Chen et al.\cite{chen2015automatic} proposed a joint learning model that combines the appearance of vertebrae and adjacent conditional dependencies of neighboring vertebrae. However, solely considering local appearance information results in a lack of overall sequence information. Similarly, Wang et al.\cite{wang2021automatic} trained a key-point localization model to jointly search for optimal identification results under the soft constraint of inter-vertebrae distance and the hard constraint of vertebrae label continuity. Payer et al.\cite{payer2020coarse} completed vertebrae localization from coarse to fine, and then regressed vertebrae thermograms using a SpatialConfiguration-Net to complete vertebrae identification. Tao et al.\cite{tao2022spine} defined vertebrae labeling as a one-to-one ensemble prediction problem and introduced a global loss that enforces the preservation of sequential relationships of vertebrae. Although it takes global sequence information into account, there is still room for improvement in the performance. Meng et al.\cite{meng2023vertebrae} proposed a CNN-based recurrent framework that utilizes prior knowledge to iteratively enforce recognition results to be consistent with statistical anatomy. However, the time required for the complex looping processing presents challenges in meeting the real-time requirements of clinical environments. To enhance recognition speed, Wu et al.\cite{wu2023multi} leveraged a pre-trained model and a multi-view contrastive learning algorithm to convert the task into a multi-view 2D vertebrae recognition task. The methods mentioned above employ different strategies to integrate global and local information, but each strategy introduces some unavoidable limitations that render them unsuitable for clinical application.

\subsection{Contrastive learning}
Recently, contrastive learning has achieved good performance in medical image tasks by enforcing that the embedded features of similar images are closed in the latent space, while dissimilar images are separated \cite{chen2020simple,he2020momentum,henaff2020data}. Iwasawa et al.\cite{iwasawa2021label} introduced contrastive learning as a subtask within a multitask segmentation model, forming a regularization branch to enhance the model's performance in medical image segmentation. Hu et al.\cite{hu2021semi} proposed a supervised local contrastive loss and applied contrastive learning to extract both global and local features from unlabeled data, resulting in successful medical image segmentation with high performance. Wang et al.\cite{wang2022few} incorporated contrastive learning into their segmentation framework to enhance feature recognition by imposing intra-class cohesion and inter-class separation between support and query features. Basak et al.\cite{basak2023semi} proposed a pixel-level contrastive learning for domain adaptation. Chen et al.\cite{chen2023ascon} proposed a MAC-Net that combines global non-contrastive and local contrastive modules, to leverage intrinsic anatomical semantic features at the patch level for enhancing low-dose CT denoising. Our method targets individual vertebra bounding box images, utilizing their labels to differentiate between positive and negative sample pairs, in order to pre-train a identification anchor network.


\subsection{Uncertainty estimation}
Modern neural networks are highly effective prediction models, but they frequently struggle to identify when their predictions may be inaccurate. As indicated in the Introduction, there are two primary forms of uncertainty: data (aleatoric) uncertainty and model (epistemic) uncertainty. Many scholars used Bayesian neural networks to estimate the uncertainty in medical image tasks. Wachinger et al.\cite{wachinger2014gaussian} proposed the utilization of Bayesian inference with Gaussian processes, wherein the covariance matrix of the posterior distribution of Gaussian processes serves to estimate the uncertainty in interpolation. Laves et al.\cite{laves2020uncertainty} introduced Bayesian approach with Monte Carlo dropout to quantify both aleatoric and epistemic uncertainty to estimate uncertainty in inverse medical imaging tasks. Likewise, there have been investigations into the use of convolutional neural networks for uncertainty estimation. Devries et al.\cite{Devries2018} proposed a method for acquiring confidence estimates for neural networks that is straightforward to implement and yields outputs that are intuitively interpretable. Liu et al.\cite{liu2020uncertainty} proposed a novel uncertainty estimation metric, called maximum active dispersion (MAD), for the estimation of image-level uncertainty in localization tasks. Qiu et al.\cite{qiu2021modal} employed a discrete set of latent variables derived from a conditional generative model. Each variable represents a latent pattern hypothesis that elucidates an input-output relationship, thereby effectively estimating uncertainty. In contrast to these methods, we weight and modify the final results by estimating the uncertainty of the classification network's prediction for each vertebrae. By leveraging the uncertainty during message fusion, we can avoid the potential biases from neighboring vertebrae which could compromise recognition performance. 

\begin{figure*}[ht]
\includegraphics[width=1\textwidth]{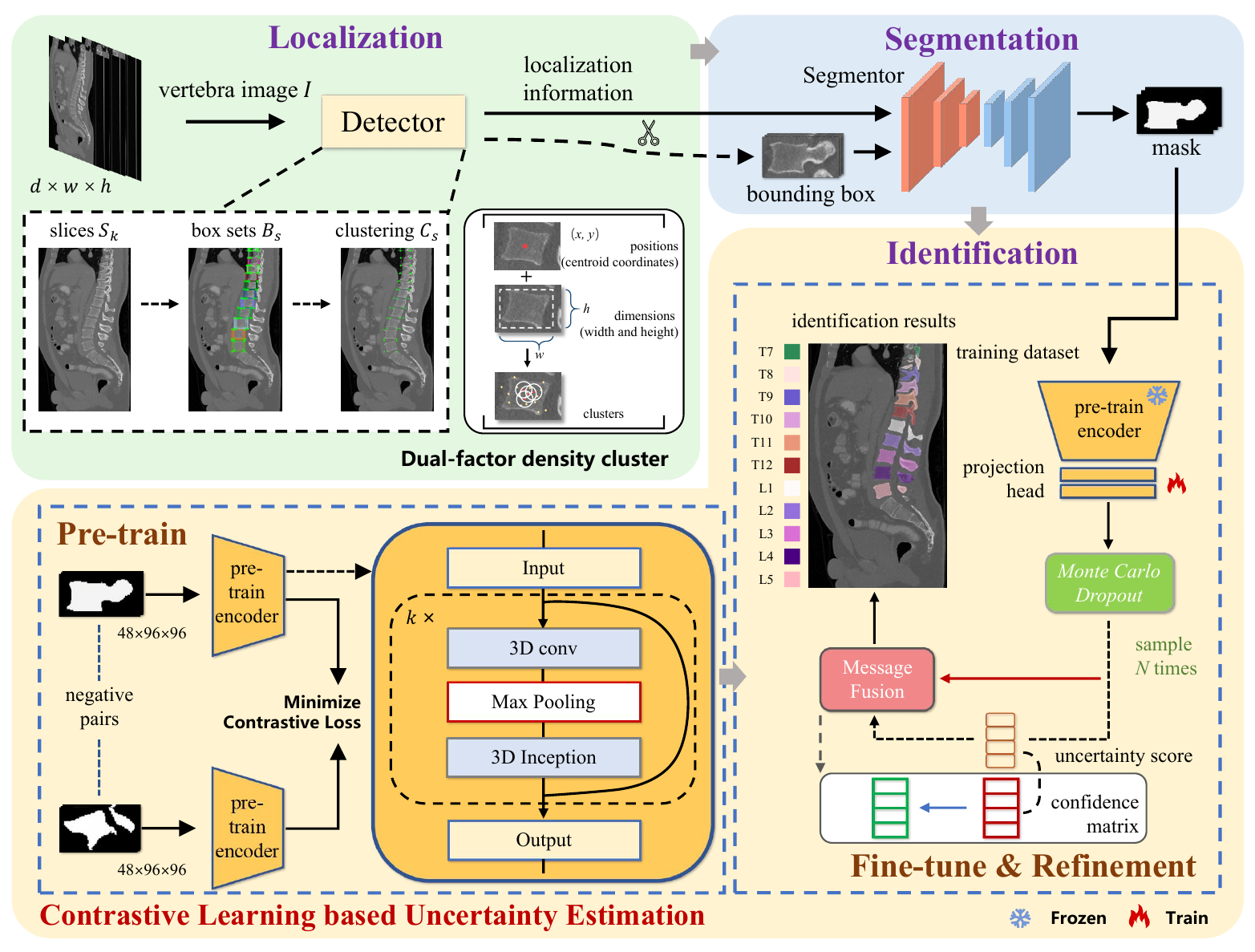}
\caption{An overview of SpineCLUE. The three consecutive stages are connected by arrows. The scissors icon indicates the use of localization information to crop finely bounding boxes suitable for segmentation to obtain segmentation masks that include transverse and spinous processes. For localization, the dashed box on the left shows the process of pre-training and the main network used. The right dashed box shows the flow of fine-tuning, uncertainty information estimation and message fusion. In this case, the spark icon indicates that the parameters are trained without freezing the gradient of the network. And the snowflake icon indicates freezing the gradient and not updating the weights of the network.}
\label{fig:main-framework}
\end{figure*}

\section{Method}
In order to address the challenge of arbitrary FOVs and better improve the performance of vertebrae identification, we propose a three-stage method as show in Fig.~\ref{fig:main-framework}, where each step will use the information extracted in the previous step for better inference. Specifically, stage 1 involves clustering the localization results to obtain the centers of vertebrae for cropping the bounding boxes of individual vertebrae segmentation. Stage 2 performs deep learning based binary segmentation on the vertebrae to generate masks that are favorable for identification. We additionally incorporated the positional information from stage 1 represented by Gaussian heatmaps into one channel of the segmentation model as our segmentation method. In stage 3, a pre-trained model obtained through contrastive learning is used for identification, simultaneously fusing the uncertainty of the model to refine the identification sequence. We use the vertebral masks obtained in stage 2 as input rather than the original image, which can make the identification network pay more attention to the shape features and vertebrae targets rather than the background.

\subsection{Dual-factor density clustering for vertebrae localization}
Abnormal vertebrae such as scoliosis and kyphosis result in unusual angles and postures of the vertebrae, and localization models trained on common datasets often fail to detect these vertebrae.
Therefore, we use a detector to predict the bounding boxes and likely centers of each vertebra in the 2D slices of 3D volumes. The vertebrae in sagittal and coronal planes can be marked with bounding boxes, thereby eliminating the influence of physiological curvature for different categories of vertebrae. We perform dual-factor clustering on the results from all slices based on two factors: dimensions (height and width) and positions (center coordinates), which do not require predefined numbers and shapes of the clusters and is able to filter out the noisy localization results in the 2D slices.

Given a preprocessed 3D CT image $I\in \mathbb R^{d\times h\times w}$, we can generate $k$ slices $I_s=\{S_1,S_2,...,S_k\}$, where $S_i\in \mathbb R^{d\times w}$. Then $I_s$ is inputted into the localization model to obtain the initial coordinates and dimensions (width and height) of detection boxes $B_s=\{b_1,b_2,...,b_k\}$ encompassing the entire vertebrae region, where $b_i=\{(x_{i,1},y_{i,1},w_{i,1},h_{i,1}), 
(x_{i,2},y_{i,2},w_{i,2},h_{i,2}),..., x_{i,l_i},y_{i,l_i},w_{i,l_i},h_{i,l_i})|l_i\leq n\}$, $n$ denotes the number of vertebrae contained in the image that can be detected and $l_i$ represents the total number of target frames for a single vertebra. To mitigate the impact of interferences present in the images on the prediction of vertebrae centers, we use a larger number for the density clustering parameter $k$. 
The neighborhood box density is defined to exclude the ngative influence of noisy boxes, with the following formula:
\begin{equation}
\label{eq:1}
\rho (b_i,\epsilon)=\frac{N_{i,\epsilon}}{l_i}
\end{equation}
where $\epsilon$ denotes the radius of the neighborhood, and $N_{i,\epsilon}$ denotes the total number of target centers contained in the 3D neighborhood centered on $b_i$ in vertebrae $i$ with $\epsilon$ as the radius.

We apply DBSCAN \cite{DBSCAN} clustering algorithm to density function $\rho (b_i,\epsilon)$ to obtain likely centers. Considering the poor quality of spine images, which may result in target boxes encompassing multiple vertebrae or incomplete coverage of individual vertebrae, we also perform density clustering on the height and width of the target boxes. This approach helps eliminate noisy boxes caused by the issue mentioned above. 

\subsection{Supervised contrastive learning for vertebrae identification}
To improve vertebrae identification performance with arbitrary FOVs and enhance the model's ability to address inter-class similarity and intra-class variability, we utilize supervised contrastive learning strategy to pre-train the model as illustrated in Fig~\ref{fig:main-framework}. The core concept of supervised contrastive learning is to minimize the distance between the positives while maximizing the distance between the negatives. We use the vertebrae labels to determine whether the samples are positive or negative and then employ a 3D ResNet-101 \cite{3DResNet} with the Inception blocks \cite{szegedy2015going} as the backbone and pre-train the model using supervised contrastive learning method. The loss function used during pre-training is as follow:
\begin{equation}\label{eq:2}
\begin{split}
\mathcal{L}_{supcon}=& \sum_{v\in V}\mathcal{L}_{supcon}^{(v)}\\
=& \sum_{v\in V}-log \Bigg\{\frac{1}{|G(v)|} \sum_{g\in G(v)} \frac{exp(z_v\cdot z_g/\tau)}{\sum_{a\in A(v)}exp(z_v \cdot z_a /\tau)} \Bigg\} 
\end{split}
\end{equation}
where, let $v\in V \equiv\{1...2T\}$ be the index of $2T$ random augmented vertebrae images. $\tau$ is a scalar temperature parameter, and $A(i)\equiv V \, \backslash \, \{v\}$. $G(v)\equiv \{p\in A(v):\Tilde{y}_g=\Tilde{y}_v\}$ is the set images of all positives in $V$ distinct from $v$. $z_v$ denotes the embeddings of the vertebrae bounding box images obtained by the encoder of backbone.

The loss function maintains the summation over negative, similar to noise contrastive estimation \cite{gutmann2010noise} and N-pair losses \cite{sohn2016improved}. By increasing more negatives, the ability to discriminate between signal and noise (negatives) is enhanced. This property is useful for resolving inter-class similarity and intra-class variability problems and has been confirmed by numerous studies \cite{he2020momentum,henaff2020data,tian2020contrastive}.

We perform two rounds of random augmentation to the input images, including random flipping, translation, rotation, and adding Gaussian noise. Subsequently, we fine-tune the additional projection head, with the gradients of the backbone frozen. When fine-tuning the model, a sequence-level enforced loss implemented by a dynamic programming algorithm is employed to impose prior knowledge constraints to the identification results. The calculation formula is expressed as:
\begin{equation}
\label{eq:3}
\begin{split}
v_{loss}&[\,i\,]=\left\{
\begin{aligned}
max(v_{loss}[i],& v_{loss}[j]+1), j<i\, and\, seq[i]>seq[j]\\
1&, \qquad\qquad\qquad\qquad\quad0<=i<n\\
\end{aligned}
\right.\\
&\qquad\qquad\qquad\quad \mathcal{L}_{se}=n-max(v_{loss})  
\end{split}
\end{equation}

Vertebrae identification involves both category prediction and sequence prediction, so mean squared error and cross-entropy loss functions which are suitable for both tasks are used in the process of fine-tuning. We weight cross-entropy $\mathcal{L}_{ce}$ and mean squared error loss functions $\mathcal{L}_{mse}$ along with $\mathcal{L}_{se}$ to form the total loss function in the fine-tuning $\mathcal{L}_{total}$:
\begin{equation}
\label{eq:4}
\mathcal{L}_{total}=\alpha \, \mathcal{L}_{se}+\beta \, \mathcal{L}_{mse}+\gamma \, \mathcal{L}_{ce}
\end{equation}

\subsection{Uncertainty estimation for identification refinement}
\subsubsection{Vertebrae uncertainty estimation}
\begin{figure}[t]
\centering
\includegraphics[width=0.4\textwidth]{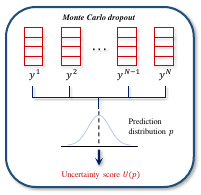}
\caption{Overview of uncertainty estimation module. The prediction distribution is estimated by sampling the confidence matrix $N$ times. Uncertainty scores are calculated by computing the entropy of the prediction distribution $p$.}
\vspace{-2pt}
\label{fig3}
\end{figure}

After fine-tuning the projection head, we complement a class uncertainty estimation module to aggregate sequential information between long-distance spines as shown in Fig.~\ref{fig3}. During training, we employ a Monte-Carlo dropout \cite{MCDropout} operation in the linear layer, deactivating a certain probability $p_i$ of neurons:
\begin{equation}
\label{eq:5}
p_i\approx \frac{1}{N} \sum_n^NSoftmax(f^{W_n}_i(x))
\end{equation}
where $N$ signifies the total number of samples, $i$ represents the index of the vertebrae, and $f^{W_n}$ denotes the neural network with parameters $W_n$. The predictions will be obtained through $N$ rounds of forward propagation guided by $p$. And the uncertainty for each vertebrae is estimated by calculating the variance and information entropy of the predictions as Eq.~\ref{eq:6}:
\begin{equation}\label{eq:6}
U(p)=-\sum_{c=1}^Cp_{i,c}\cdot logp_{i,c}
\end{equation}

During inference, the dropout activation is maintained, and predictions are also sampled $N$ times to calculate their uncertainty score.

\subsubsection{Uncertainty message fusion}
\begin{figure}[t]
\centering
\includegraphics[width=1\textwidth]{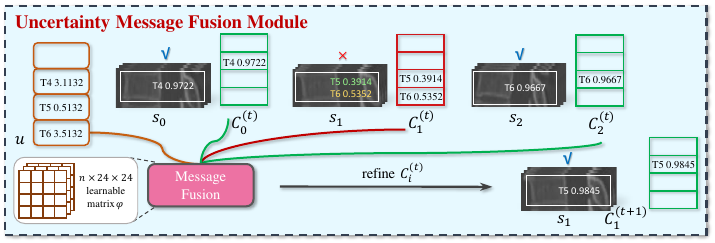}
\caption{Architecture of the uncertainty message fusion module. The uncertainty message fusion module utilizes the confidence matrices $(C_0^{t},\;C_2^{t})$ and uncertainty information $u$ of the correctly identified vertebrae $s_0$ and $s_2$ to correct the pre-identification results of $s_1$.}
\label{fig:message-fusion}
\end{figure}

Only using the network after supervised contrastive learning pre-training and fine-tuning for vertebrae identification can not pay attention to the adjacent vertebrae, resulting in individual prediction being mis-classified into neighboring vertebrae. Therefore, we design an uncertainty message fusion module to rectify erroneous pre-recognition confidence scores as shown in Fig.\ref{fig:message-fusion}. The essence of message fusion lies in adjusting the confidence of the current vertebrae by incorporating the confidence and uncertainty scores of neighboring vertebrae. A learnable matrix is used to aid in message fusion, ensuring that incorrect confidence scores are iteratively corrected, ultimately converging to their maximum values corresponding to the correct labels. The fusion process is described by Eq.~\ref{eq:7}, which illustrates how vertebrae aggregate neighboring message.
\begin{equation}\label{eq:7}
\begin{split}
\mathcal{C}_i^{t+1} =\; &\frac{1}{\lambda}(\mathcal{C}_i^{t}+\theta \cdot \omega_i^t)\\
\omega_i^t =\sum_{j\in Neighbor(i)}&\frac{1}{dis(i,\,j)}\times u_j\times \mathcal{C}_j^t\times \varphi_j
\end{split}
\end{equation}

where: (1) In the process of message fusion, we use hops to indicate the fusion times. The $t+1$ hop confidence $\mathcal{C}_i^{t+1}$ will iteratively absorb the values from $t$ hop $\mathcal{C}_i^{t}$ and the message transmitted by all surrounding neighboring nodes $j$, where $j\in Neighbor(i)$. Specifically, the message is computed through the wise product of uncertainty scores $u_j$, confidence from $t$ hop $\mathcal{C}_j^t$, and learnable matrix $\varphi_j$ (each neighbor corresponds to a 24 $\times$24 trainable parameters). Additionally, the distance along the vertebrae $i$ and $j$ $dis(i,\,j)$ is used as a weight. (2) Neighbor message will be aggregated onto $\mathcal{C}_i^{t}$ using a hyper-parameter $\theta$. (3) $\lambda$ is a normalization constant to force the sum of $\mathcal{C}_i^{t+1}$ to be equal to 1.

The $t$ hop vertebrae confidence $\mathcal{C}_i^{t}$ will aggregate message $\omega_i^t$ sent from the neighbors through weighted sum. To suppress irrelevant information contained in distant neighbors, we weight the messages from different neighbors based on the distance. Additionally, each vertebra has a specific learnable matrix for feature extraction, denoted by the subscript $j$ of $\varphi$.

\section{Experiments and results}
\subsection{Datasets and evaluation metrics}
We evaluated our method on two challenging public dataset: MICCAI VerSe19 and VerSe20 \cite{sekuboyina2021verse}. VerSe19 contains of 160 spinal CT with ground truth annotations. As the public challenge settings, 80 scans are used as training dataset, 40 scans for testing dataset, and 40 scans for the hidden dataset. VerSe20 contains of 319 spinal CT with ground truth annotations. Of these scans, 113 scans for training dataset, 103 scans for testing dataset, and 103 scans for hidden dataset. We divided the training dataset into two parts at 9:1 for model training and testing, respectively. And we used testing and hidden dataset as our two validation sets with the same setting as Wu et al.\cite{wu2023multi}, Tao et al.\cite{tao2022spine}, and Meng et al.\cite{meng2023vertebrae}.

The input CT images in VerSe dataset are with arbitrary FOVs and resolution. The volumes are re-sampled into an isotropic resolution of $1\times 1\times 1mm^3$ and re-oriented into the same anatomic orientation as pre-processing. During localization, the slices are resized to a size of $640\times640$. Subsequently, we will use the localization results to crop the target vertebrae into bounding boxes of size $128\times128\times128$. To reduce the parameter and training duration of the identification model, the bounding boxes in identification will be finely cropped to $48\times96\times96$ based on the mask. The final result will be transformed back to the original orientation and resolution.

Furthermore, to validate the generalization ability of SpineCLUE to a wider range of spondylosis cases, we additionally collected an abnormal dataset, comprising 62 cervical and thoracic FOV images, and 132 lumbar FOV images. The data primarily originates from two large spine datasets, Spine1K \cite{deng2021ctspine1k} and CSI2014 The dataset includes conditions such as scoliosis, degenerative deformities and metal implants (mostly in the lumbar images). Each image has a spacing from 0.45 to 5.0, and we applied the same preprocessing methods as used for the VerSe dataset.

To evaluate the performance of our method, we use the identification rate (ID-rate) and MSE as an evaluation metric. The ID-rate is defined as the ratio of correctly identified vertebrae to the total number of vertebrae. MSE represents the mean squared error between the predicted labels and the true labels.

\subsection{Implementation details}
The three-stage framework for vertebrae localization, segmentation and identification was implemented using PyTorch and each stage was separately trained on two NVIDIA GeForce RTX 3090 GPUs.

\subsubsection{Vertebrae localization}
In the localization, we employed YOLOv7-X \cite{yolov7} as our detector. We sliced 20 CT images from the training set and manually annotated vertebrae bounding boxes for model training. During model training, the learning rate was set to 1e-3. The training process used batchsize of 64 for 500 epochs until reaching the early stop criteria. For model inference, the number of slices $k$ on both sagittal and coronal planes was set to 200.

\subsubsection{Vertebrae segmentation}
The network used in vertebrae segmentation is TransUNet \cite{2021TransUNet}. The initial learning rate was set to 1e-4 and apply Adam optimizer with a 1e-4 weight decay schedule. We conducted training for a total of 300 epochs, utilizing a batch size of 32.

\subsubsection{Vertebrae identification}
For the identification, we applied a learning rate of 1e-2 for the pre-trained 3D ResNet-101 backbone and a learning rate of 1e-5 for fine-tuning the projection head respectively. Their batchsizes were 32 and 64, respectively. The models underwent pre-training for a total of 500 epochs, followed by fine-tuning for an additional 100 epochs. The number of uncertainty estimation sampling $N$ was set to 20, and the weight of message fusion $\theta$ in Eq.\ref{eq:7} was set to 0.1. Regarding the hyper-parameters for the loss function, we set $\alpha$, $\beta$, $\gamma$ in Eq.\ref{eq:4} to 0.1, 0.5, 1 respectively. 

\begin{table*}[t]
\centering
\caption{Comparative experimental results on VerSe19 and VerSe20 datasets. The ID-rate metric is utilized to express the percentage value of correctly identified results in relation to the total number of identified vertebrae. The outcomes of our method are denoted by the use of bold formatting, while the most significant results are emphasized through underlining. Dashes in the table indicate that the experimental results are not available because the code is not open source, etc.}
\label{tab:comparision}
\resizebox{0.99\textwidth}{!}{
\begin{tabular}{ccccc}
\hline
& \bfseries VerSe19 public set & \bfseries VerSe19 hidden set & \bfseries VerSe20 public set & \bfseries VerSe20 hidden set\\
\hline
& ID-rate (\%) & ID-rate (\%) & ID-rate (\%) & ID-rate (\%) \\
\hline
Chen et al.\cite{chen2020hmm} & 95.61 & 96.58 & 96.94 & 86.73 \\
Payer et al.\cite{payer2020coarse}& 95.06 & 92.82 & 95.65 & 94.25 \\
Tao et al.\cite{tao2022spine} & - & - & 97.22 & 96.74 \\
Meng et al.\cite{meng2023vertebrae} & 96.59 & 95.38 & 96.45 & 95.15 \\
Wu et al.\cite{wu2023multi} & - & - & 98.12 & 96.45 \\
\hline
\bfseries Ours & \underline{\bfseries97.92} & \underline{\bfseries98.14} & \underline{\bfseries 98.73} & \underline{\bfseries 97.87} \\
\hline
\end{tabular}
}
\end{table*}

\begin{figure*}[t]
\centering
\subfigure[]{
    \includegraphics[width=0.23\textwidth]{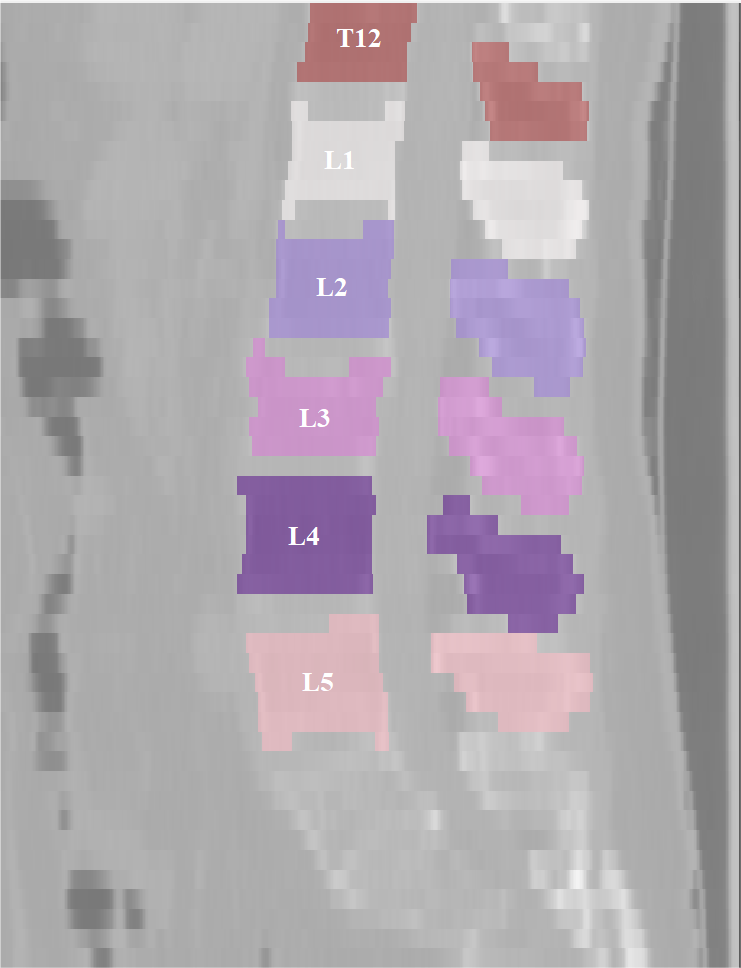}
}
\subfigure[]{
    \includegraphics[width=0.23\textwidth]{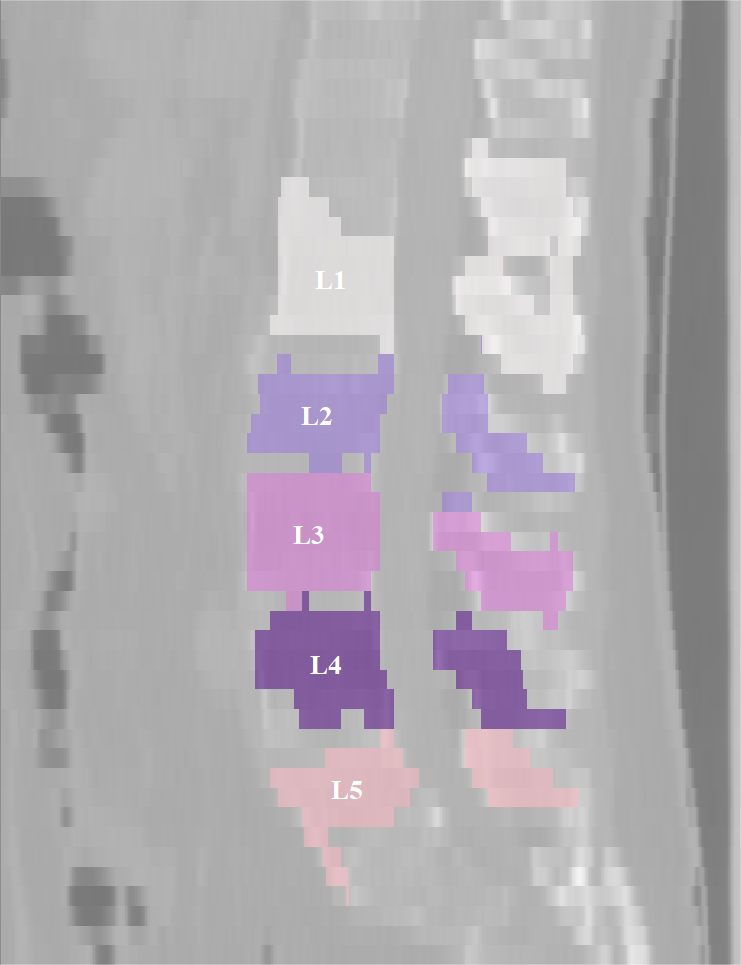}
}\subfigure[]{
    \includegraphics[width=0.23\textwidth]{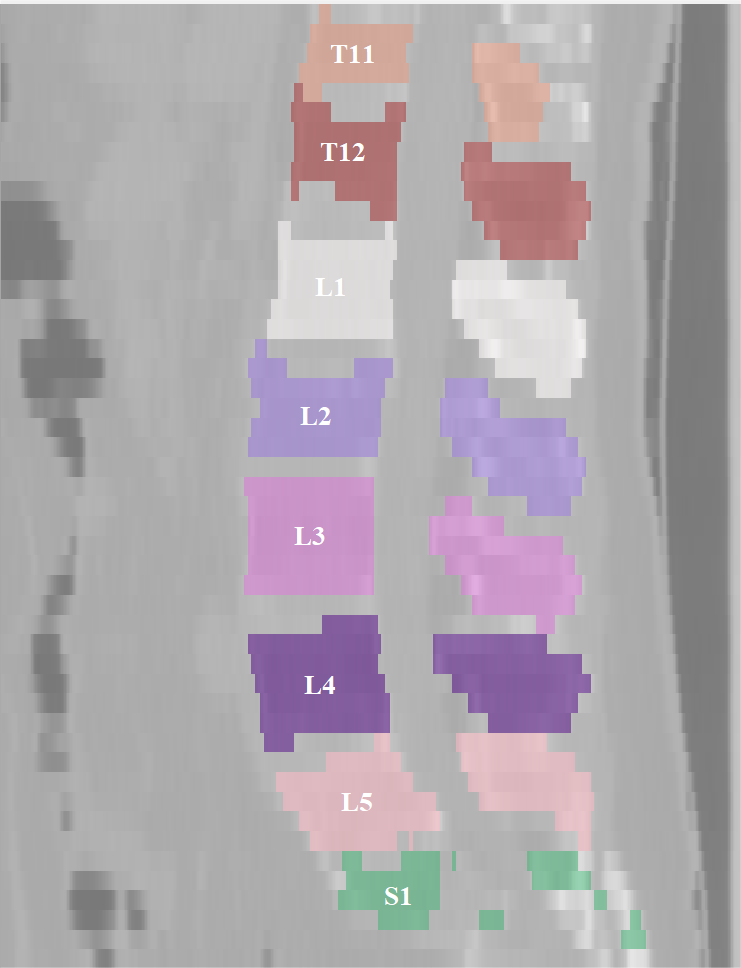}
}\subfigure[]{
    \includegraphics[width=0.23\textwidth]{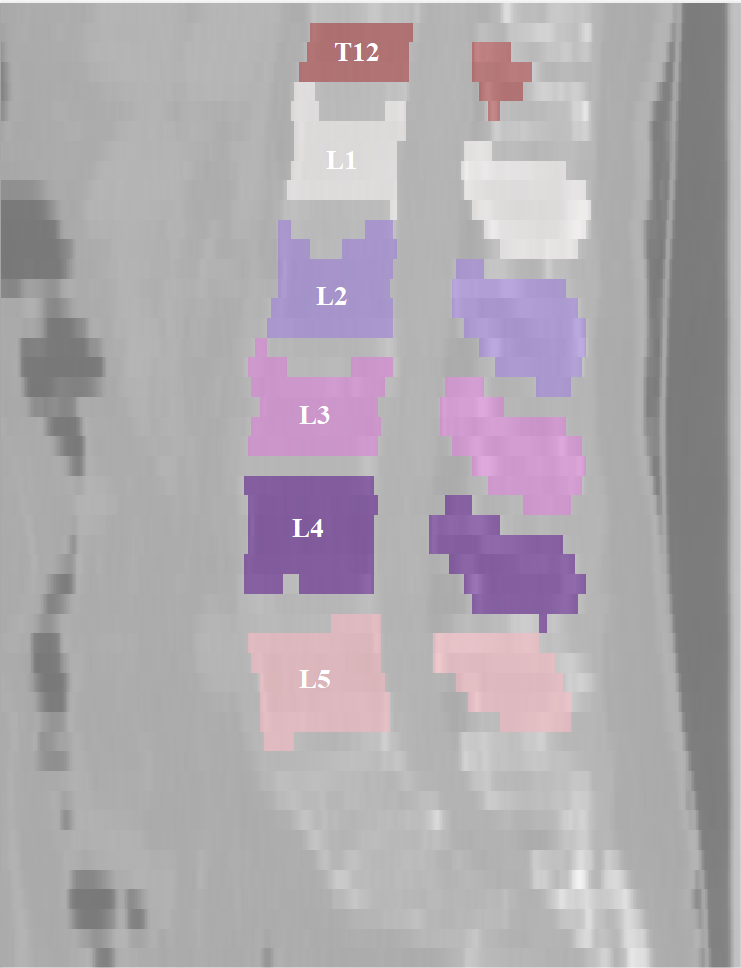}
}
\caption{Qualitative results of low-resolution CT in sagittal plane. The pictures are sagittal slices of spine covered with the corresponding vertebrae masks, and the four sub-figures labeled as a, b, c, d in the above figure represent: ground truth, Payer et al.\cite{payer2020coarse}, Meng et al.\cite{meng2023vertebrae}, and SpineCLUE respectively. Vertebrae masks are indicated using different colors and are annotated on the images. In this subjected image, the range of the ground truth is from T12 to L5, excluding the sacrum (S1).}
\label{fig:visualization low-resolution comparasion}
\end{figure*}

\begin{figure}[!t]
\centering
\includegraphics[width=0.51\linewidth]{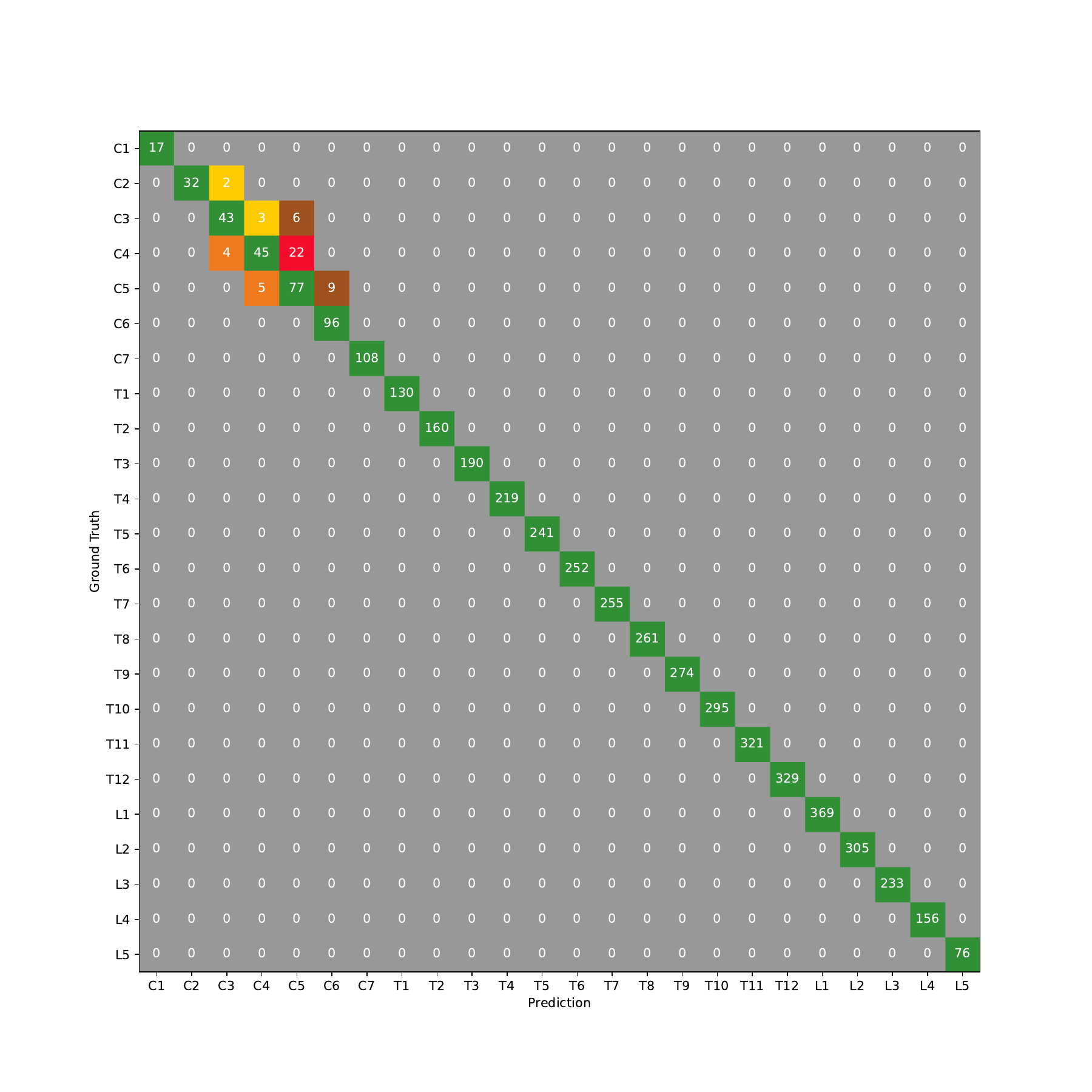}
\hspace{-5mm}
\includegraphics[width=0.51\linewidth]{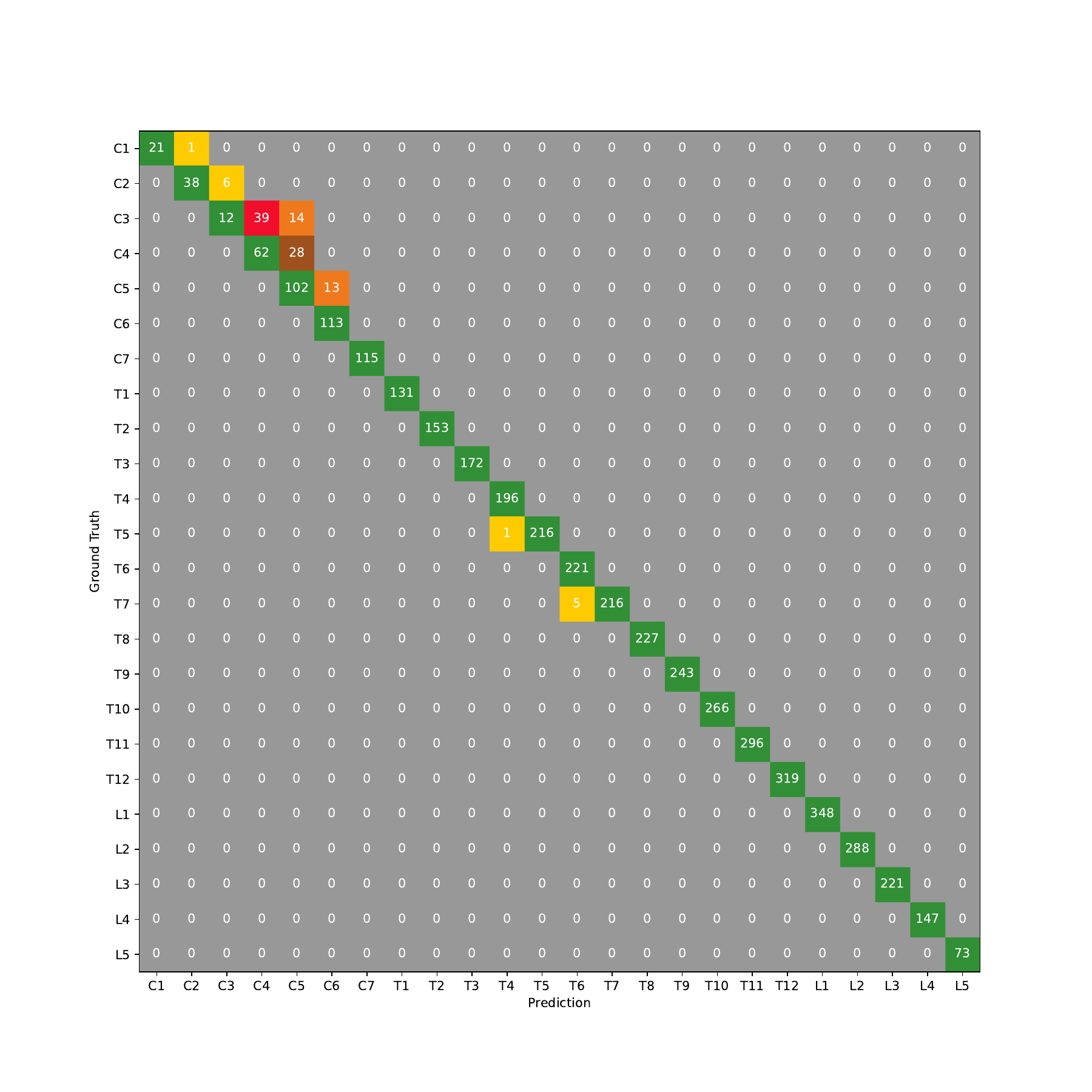}
\caption{The confusion matrix for the VerSe20 dataset. The results from the public set are presented on the left, while the results from the hidden set are displayed on the right. The x and y axes are used to list the 24 vertebrae categories in order. Green blocks positioned along the diagonal represent accurately identified outcomes, whereas other outcomes that have been incorrectly identified are depicted with increasingly darker shades as the numerical value increases.}
\label{fig:confusion-matrix}
\end{figure}

\subsection{Comparison with existing methods}
We compared SpineCLUE with five methods, as shown in Table~\ref{tab:comparision}. SpineCLUE has demonstrated outstanding performance on both the VerSe19 and VerSe20 datasets. It achieved the highest scores among all competing models. Specifically, the ID-rates of SpineCLUE reached 97.92\% and 98.14\% on the public and hidden sets of VerSe19. Similarly, on the public and hidden sets of VerSe20, SpineCLUE achieved ID-rates of 98.73\% and 97.87\%, respectively.

Most methods exhibit fluctuations in the identification scores for hidden and test sets, mainly due to the different distribution of abnormal data in these two datasets. Specifically, the spine-level method can produce abnormal results for the entire spine, often due to partial vertebrae misses caused by compression fractures, poor image quality, and metal artifacts. This reflected the fact that the vertebrae-level identification paradigm of SpineCLUE is more superior in terms of robustness compared to other spine-level methods.

Qualitative results of our method and the baseline methods on the low-resolution image cases are show in Fig.\ref{fig:visualization low-resolution comparasion}. Payer et al.\cite{payer2020coarse} failed to accurately identify the T12 vertebra, which led to its absence in the final results. Moreover, there is a misclassification issue where the L1 and L2 vertebrae are merged together and identified as L1. As a result, there is a one-category deviation in the overall identification result when compared with the ground truth. Meng et al.\cite{meng2023vertebrae} has a very competitive performance in localization and segmentation. It even segmented the sacrum that was not included in the ground truth, but the overall categories were also offset by one category in the final identification results. SpineCLUE demonstrated successful localization and segmentation of the vertebrae, producing accurate identification results. Nonetheless, when compared to Meng et al.\cite{meng2023vertebrae}, the segmentation results still need to be improved.

In order to gain a clearer understanding of SpineCLUE's performance, we plot the confusion matrix for the identification results of VerSe20 as shown in Fig.~\ref{fig:confusion-matrix}. The misidentification results did not surpass two categories from the correct ones, and the vast majority of thoracic and lumbar vertebrae were identified correctly after the refinement process. We also analyzed the identification results for three distinct types of vertebrae, as shown in Table~\ref{tab:different-vertebrae}. Similarly, this table provided evidence that SpineCLUE exhibits high stability, particularly in addressing the challenges of intra-class similarity and inter-class variability, which were more pronounced in the lumbar and thoracic vertebrae sections.

\begin{table}[t]
\caption{Validation results pertaining to various vertebrae categories on the VerSe20 dataset. The IDrate is displayed for cervical, thoracic, lumbar, and average conditions, respectively.}
\label{tab:different-vertebrae}
\begin{center}
\begin{tabular}{c|c|c}
\hline
& \bfseries VerSe20 public set & \bfseries VerSe20 hidden set \\
\hline
& ID-rate (\%) & ID-rate (\%) \\
\hline
Cervical & 89.13 & 86.57 \\
Thoracic & 100.00 & 99.70 \\
Lumbar & 100.00 & 100.00 \\
\hline  
\bfseries Average & \bfseries 98.73 & \bfseries 97.87 \\
\hline     
\end{tabular}
\end{center}
\end{table}

\begin{table*}[t]
\begin{center}
\caption{Ablation study results of key components. Baseline, mask, contrastive learning pre-training(CL) and message fusion (MF) indicate using complete ResNet-101 solely, using segmentation masks as input, adding contrastive learning to pre-train and fine-tune model, and incorporating uncertainty estimation and message fusion modules.}
\label{tab:ablation}
\resizebox{0.9\textwidth}{!}{
\begin{tabular}{cccc|c|c|c|c}
\hline
& & & & \multicolumn{2}{|c|}{\bfseries \;VerSe20 public set\;} & \multicolumn{2}{|c}{\bfseries \;VerSe20 hidden set\;} \\
\hline
\bfseries \;Baseline\; & \bfseries Mask\; & \bfseries CL\; & \bfseries MF\; & ID-rate (\%) & MSE & ID-rate (\%) & MSE \\
\hline
\multicolumn{1}{c}{\checkmark} & & & & 64.42 & 3.0245 & 65.66 & 5.6622 \\
\multicolumn{1}{c}{\checkmark} & \multicolumn{1}{c}{\checkmark} & & & 71.16 & 0.5455 & 70.94 & 0.6780 \\
\multicolumn{1}{c}{\checkmark} & \multicolumn{1}{c}{\checkmark} & \multicolumn{1}{c}{\checkmark} & & 94.41 & 0.2540 & 92.15 & 0.4282 \\
\multicolumn{1}{c}{\checkmark} & \multicolumn{1}{c}{\checkmark} & \multicolumn{1}{c}{\checkmark} & \multicolumn{1}{c|}{\checkmark} & \underline{\bfseries98.73} & \underline{\bfseries0.1296} & \underline{\bfseries97.87} & \underline{\bfseries0.3099} \\
\hline
\end{tabular}
}
\end{center}
\end{table*}

\subsection{Ablation study}
As shown in Table~\ref{tab:ablation}, we conduct ablation study on SpineCLUE to validate the effectiveness of key components. We established four sequentialuly added module options: 1) Baseline: utilizing ResNet-101 as the classification network for the specific task of vertebrae identification without decomposition. 2) Mask: using vertebrae masks as input to the identification network to affirm the significance of the segmentation process. 3) CL: pre-training the backbone using supervised contrastive learning and then validating the performance after fine-tuning. 4) ML: with other modules keeping active, adding uncertainty estimation and message fusion to refine the identification results. An additional metric MSE is used to highlight the effectiveness of each module from a sequence optimization perspective. Using masks as the input for identification improves both the performance of ID-rate and MSE. This demonstrates that masks can provide clearer vertebrae boundaries and morphological information for the identification model. The pre-training strategy significantly improved the ID-rate by 23.25\% and 26.49\%, and MSE by 0.2915 and 0.2498 on two test sets respectively. This validates that the vertebrae contrastive learning method successfully overcomes the challenges of identification. Since the confidence matrix obtained from the identification is accurate, our uncertainty estimation and message fusion module can effectively extract uncertainty information and use it to refine identification results. With the integration of all modules, our method ultimately achieved scores of 98.78\% and 97.87\% on ID-rate and 0.1296 and 0.3099 on MSE.

\begin{figure}[!t]
    \centering
    \subfigure[Hops $t$ in message fusion.]{
        \includegraphics[width=0.45\textwidth]{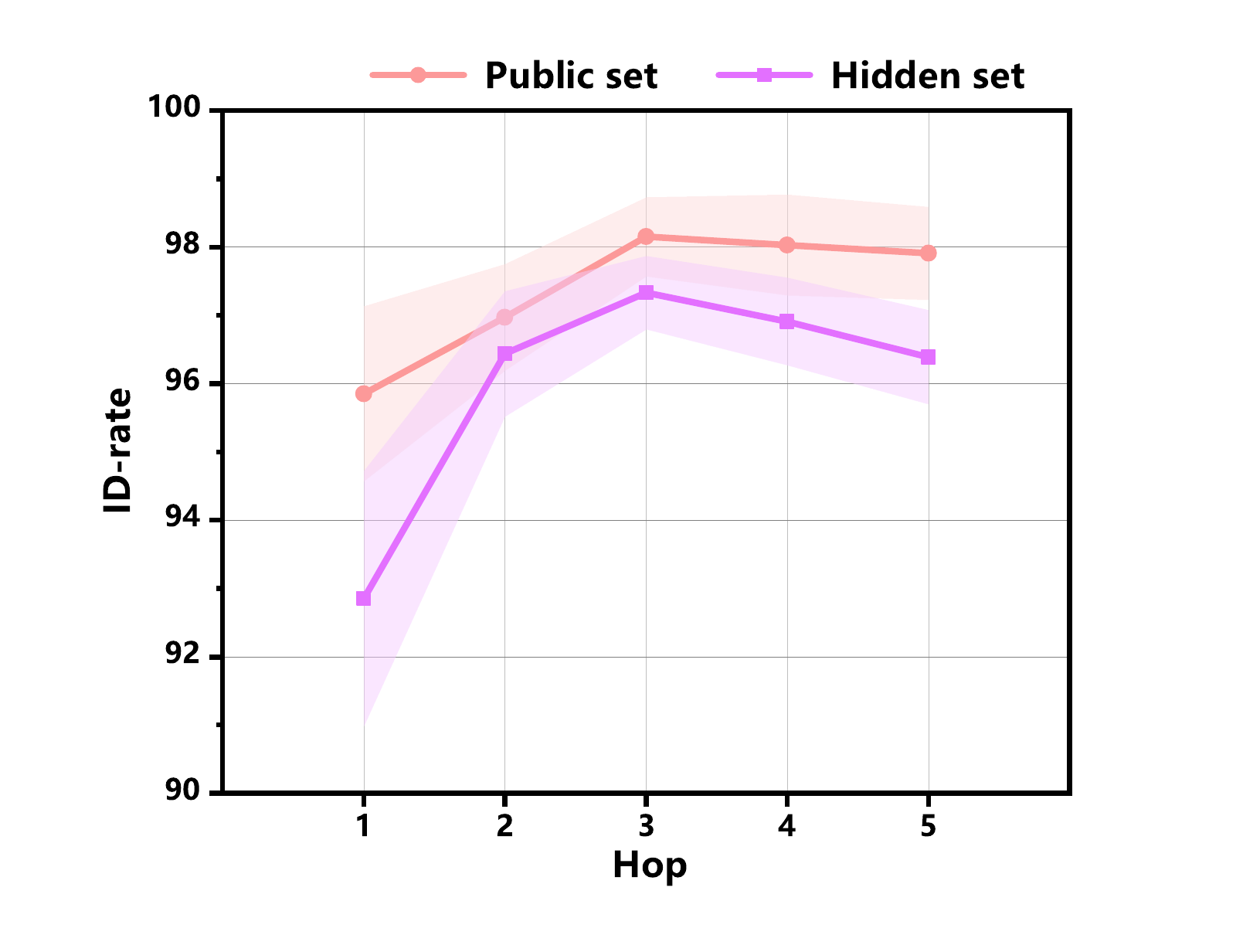}
        \label{fig:p-hop}
    }
    \subfigure[Sample times $N$.]{
        \includegraphics[width=0.45\textwidth]{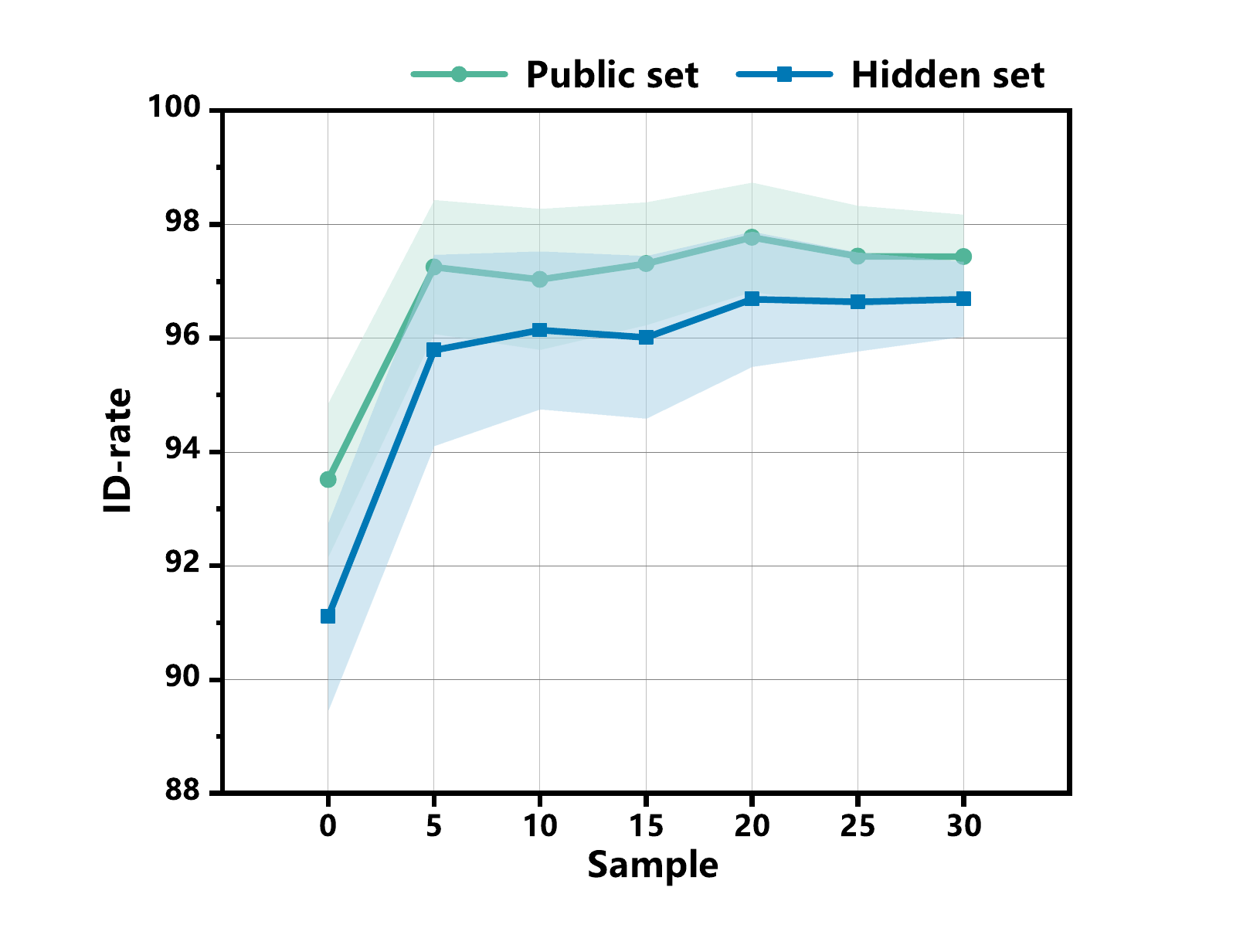}
        \label{fig:p-sample}
    }
    \subfigure[Fusion weights $\theta$.]{
        \includegraphics[width=0.45\textwidth]{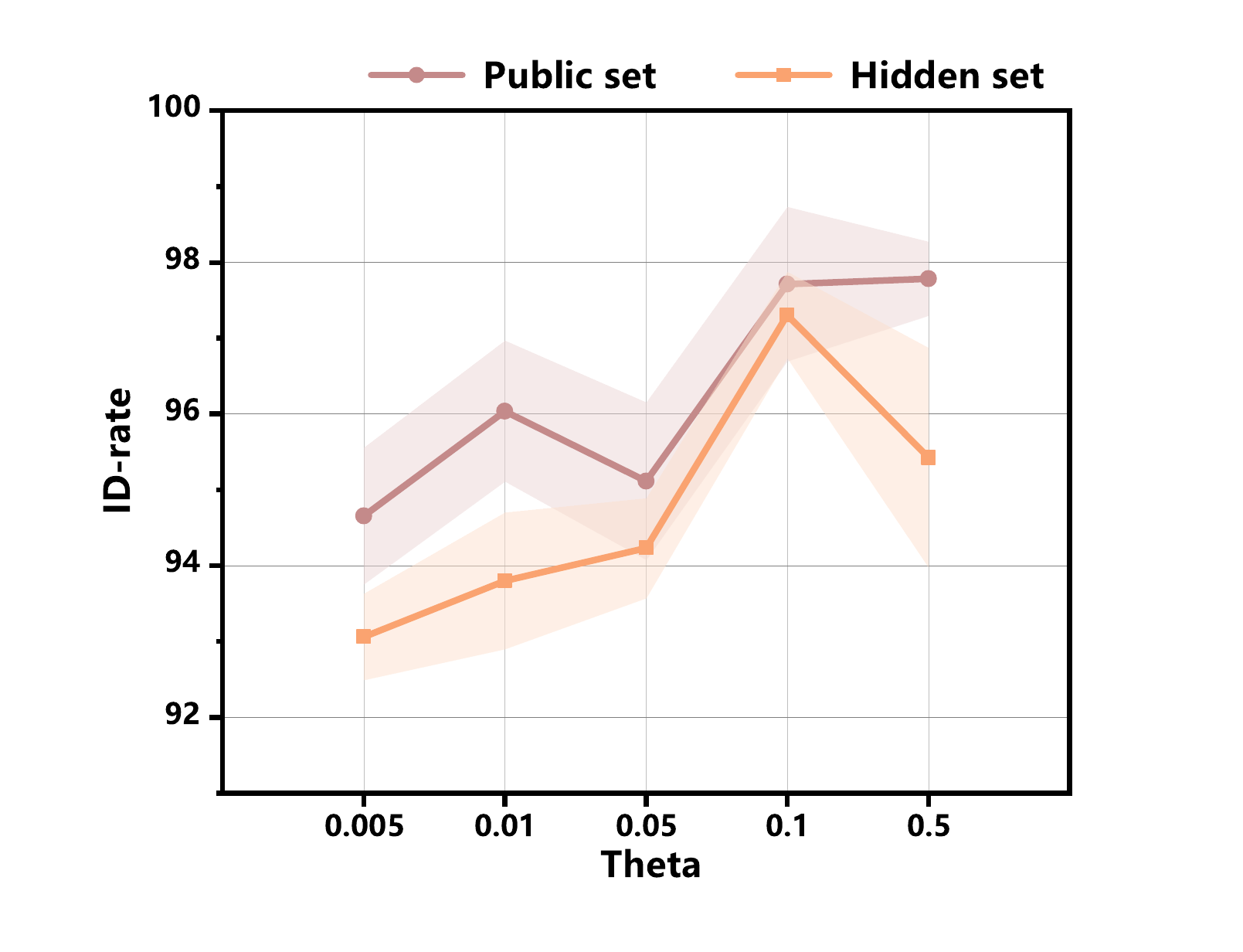}
        \label{fig:p-theta}
    }
    \subfigure[Number of neighbors.]{
        \includegraphics[width=0.45\textwidth]{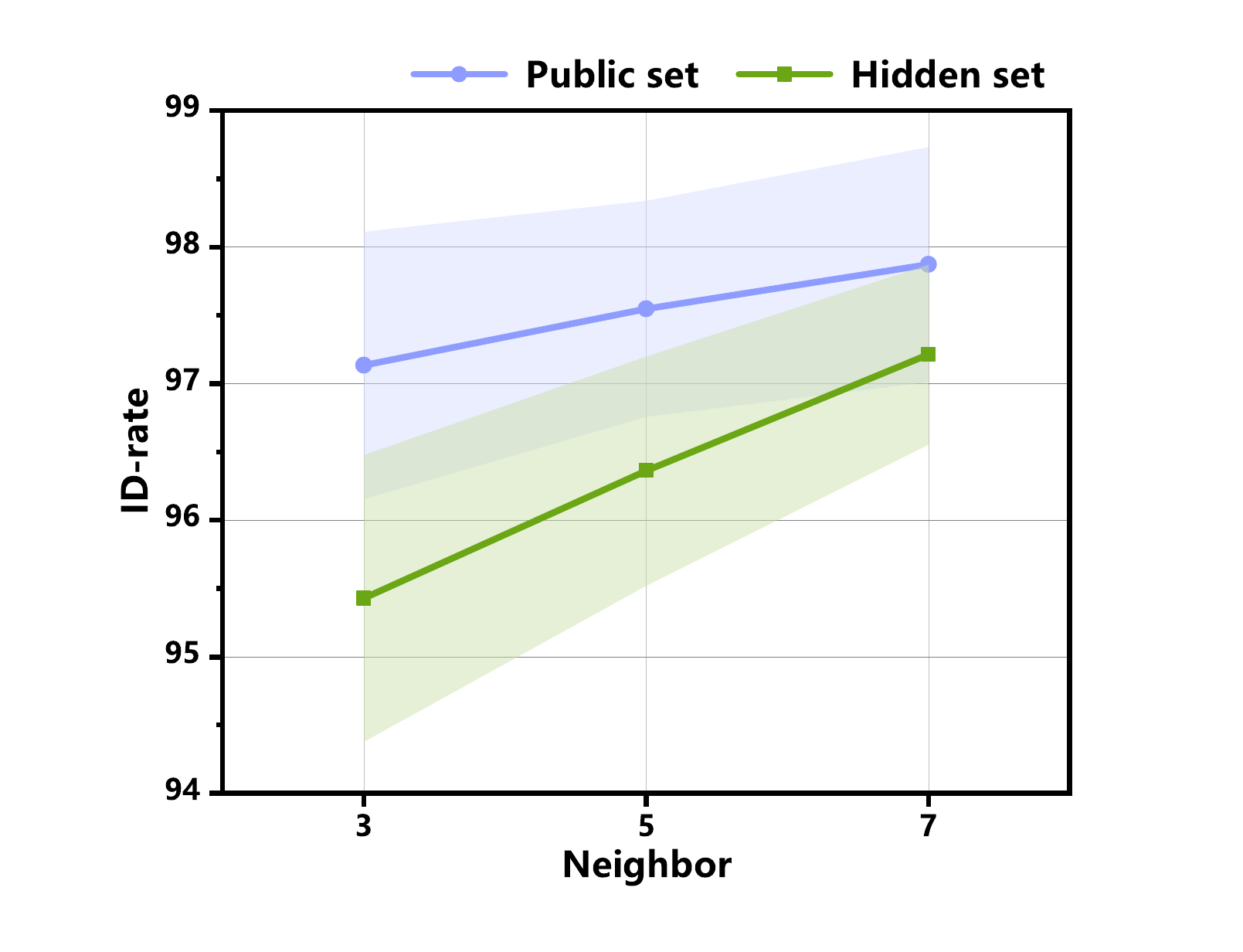}
        \label{fig:p-neighbor}
    }
    \caption{Line graphs of parameters analysis experiments. Each parameter was experimented on both the public and hidden sets of VerSe20, with the predicted ID-rate score as the dependent variable on the y-axis. The solid line represents the averaged results from multiple experiments, while the light background denotes the error interval.}
    \label{fig:parameters}
    \vspace{-2pt}
\end{figure}

\subsection{Parameters analysis}
We performed parametric analysis experiments on both the public and hidden sets of VerSe20 dataset. Hops $t$ in Eq.\ref{eq:7}, sample times $N$ in Eq.\ref{eq:5}, fusion weights $\theta$ in Eq.\ref{eq:7}, and the number of neighbors for message fusion were analyzed individually as shown in Fig.\ref{fig:parameters}. For each parameter experiment, the best performing values were utilized for the other non-primary parameters.

\subsubsection{Hops in message fusion}
The number of hops $t$ indicates the fusion level of neighboring information. As depicted in Fig.\ref{fig:p-hop}, the ID-rate of SpineCLUE increases with a greater number of hops in the beginning. When the number of hops exceeds 3, there exists a decline in the performance of SpineCLUE and the model becomes less stable. The neighbor information contains noise, and the level of noise present in the current confidence level will escalate with each aggregation. Therefore, it is not recommended to use a higher (greater than 5) number of hops. The model exhibits greater stability at hop=3, and its performance diminishes beyond this value.

\subsubsection{Sample times in uncertainty estimation}
In general, a greater number of samples $N$ will conform to a more realistic sample distribution. As depicted in Fig.\ref{fig:p-sample}, the model's stability increases with a higher number of samples. Sampling time of 20 yields optimal performance on both datasets, while higher numbers may result in a smoother sample distribution that does not effectively accommodate abnormal data. Moreover, an increased sampling time results in additional computational overhead.

\subsubsection{Fusion weights in message fusion}
In message fusion, fusion weights $\theta$ are used to regulate the fusion ratio between the neighboring messages and the original message. Lower fusion weights (<=0.05) do not significantly enhance the model as shown in Fig.\ref{fig:p-theta}. A fusion weight around 0.1 can weigh the fusion scale and partially rectify the erroneous portion of the initial message. While a larger $\theta$ will cause the model's performance degradation and instability. This mainly due to the loss of the original valid information resulting from the excessive aggregation of neighbor information. Meanwhile, the inaccurate portion of the neighbor information that cannot be mitigated by the uncertainty score may lead to interference. This phenomenon is apparent in the hidden set, where a significant decline in performance is observed.

\subsubsection{Number of neighbors in message fusion}
The number of neighbors (including target vertebrae) defines the number of vertebrae that will be assimilated by the target vertebra during the process of message fusion. The parameter is set to an odd number to ensure that each target vertebra can equally absorb vertebrae messages from both the front and back. As the majority of lumbar CT scans typically encompass only 5-8 vertebrae, we set the upper limit of the parameter at 7 to prevent any issues with performing the calculation. As depicted in Fig.\ref{fig:p-neighbor}, the ID-rate of model demonstrates a notable improvement as the number of neighbors increases, suggesting that the incorporation of more global information leads to better correction of the recognition results.


\begin{figure*}[t]
\centering
\includegraphics[width=0.245\textwidth]{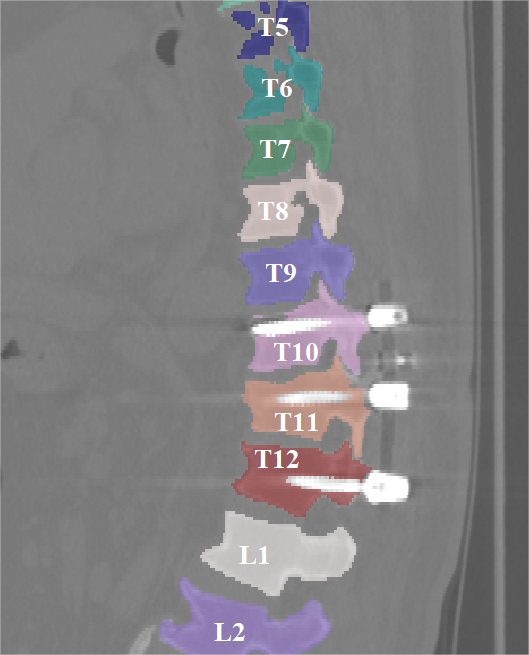}
\hspace{-2mm}
\includegraphics[width=0.245\textwidth]{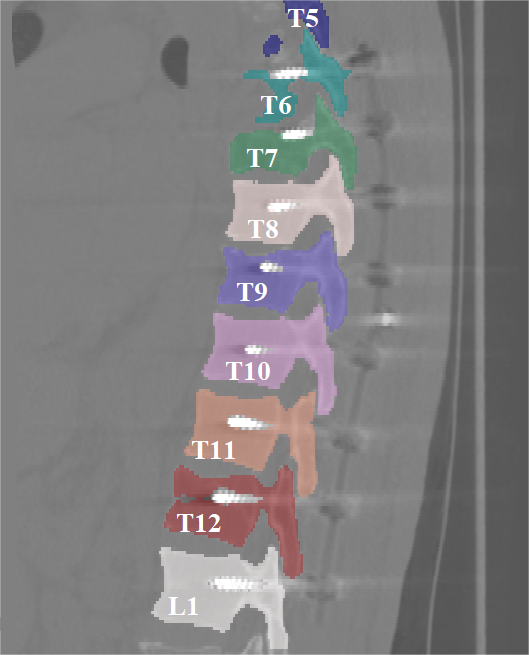}
\hspace{-2mm}
\includegraphics[width=0.245\textwidth]{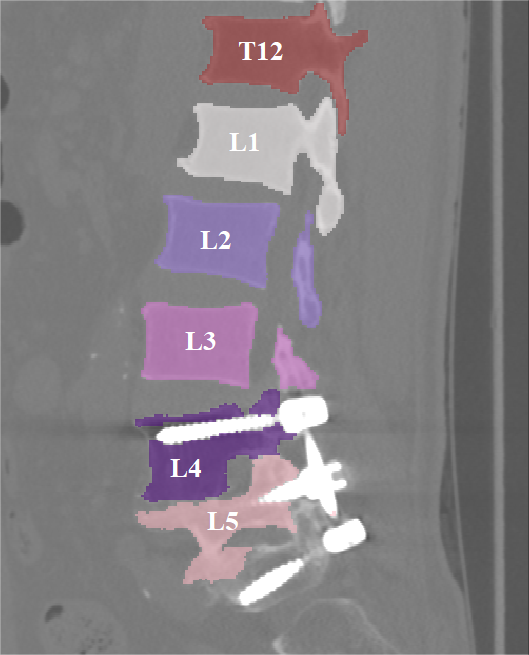}
\hspace{-2mm}
\includegraphics[width=0.245\textwidth]{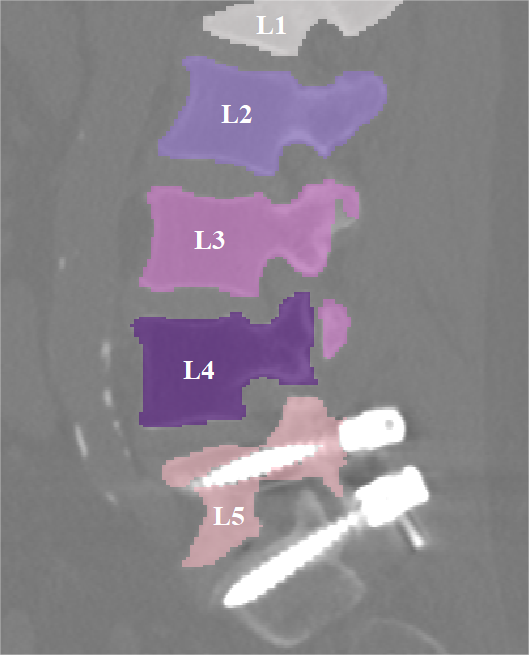}
\caption{Four experimental results of SpineCLUE for images that included metal implants. The white artifacts in the images are steel nails implanted into the spine. These nails often have HU values that deviate significantly from those of the surrounding bone and paravertebral tissues, thereby posing challenges in accurately identifying the vertebrae bodies.}
\label{fig:metal-interference}
\end{figure*}

\begin{figure*}[t]
\centering
\includegraphics[width=0.96\textwidth]{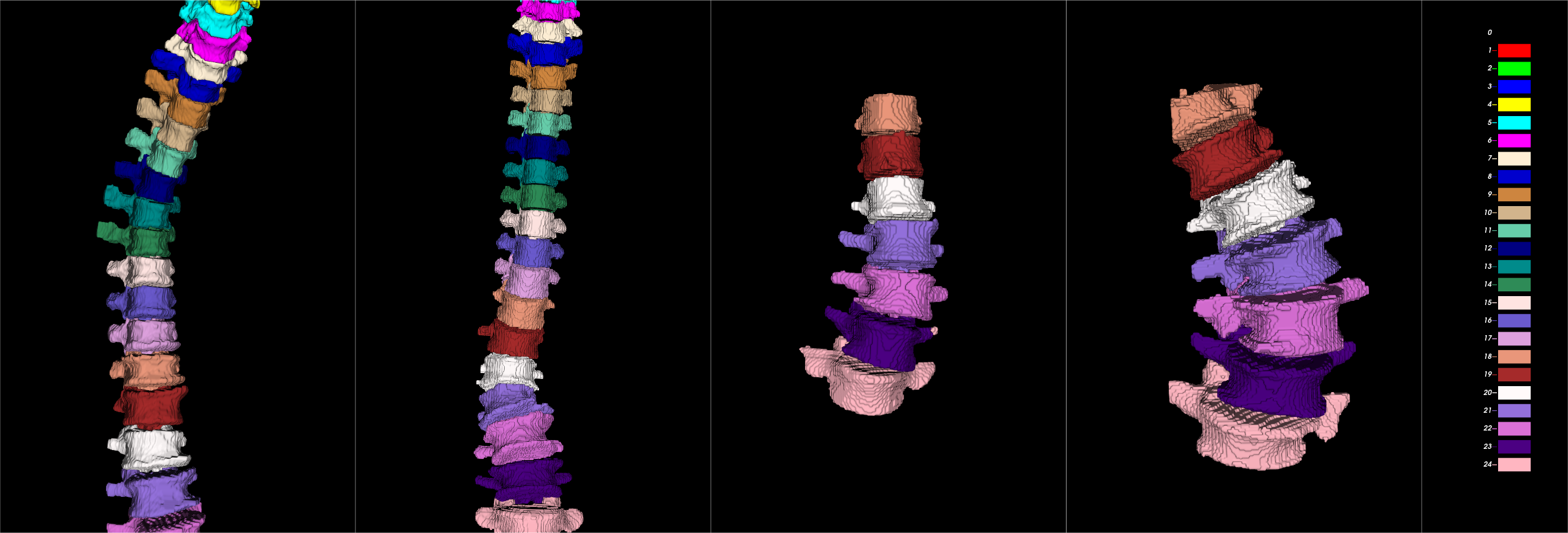}
\caption{3D schematic depicting the results of scoliosis image recognition. CTs of the thoracic, and lumbar spine are presented from the coronal perspective, respectively. Each spine shows varying locations and degrees of scoliosis. The color block diagrams on the right side depict the distinct categories of the thoracic, lumbar, and cervical spine in sequential order.}
\label{fig:bend-whole}
\end{figure*}

\subsection{Generalization evaluation on the abnormal dataset}
\begin{figure*}[t]
\centering
\includegraphics[width=0.97\textwidth]{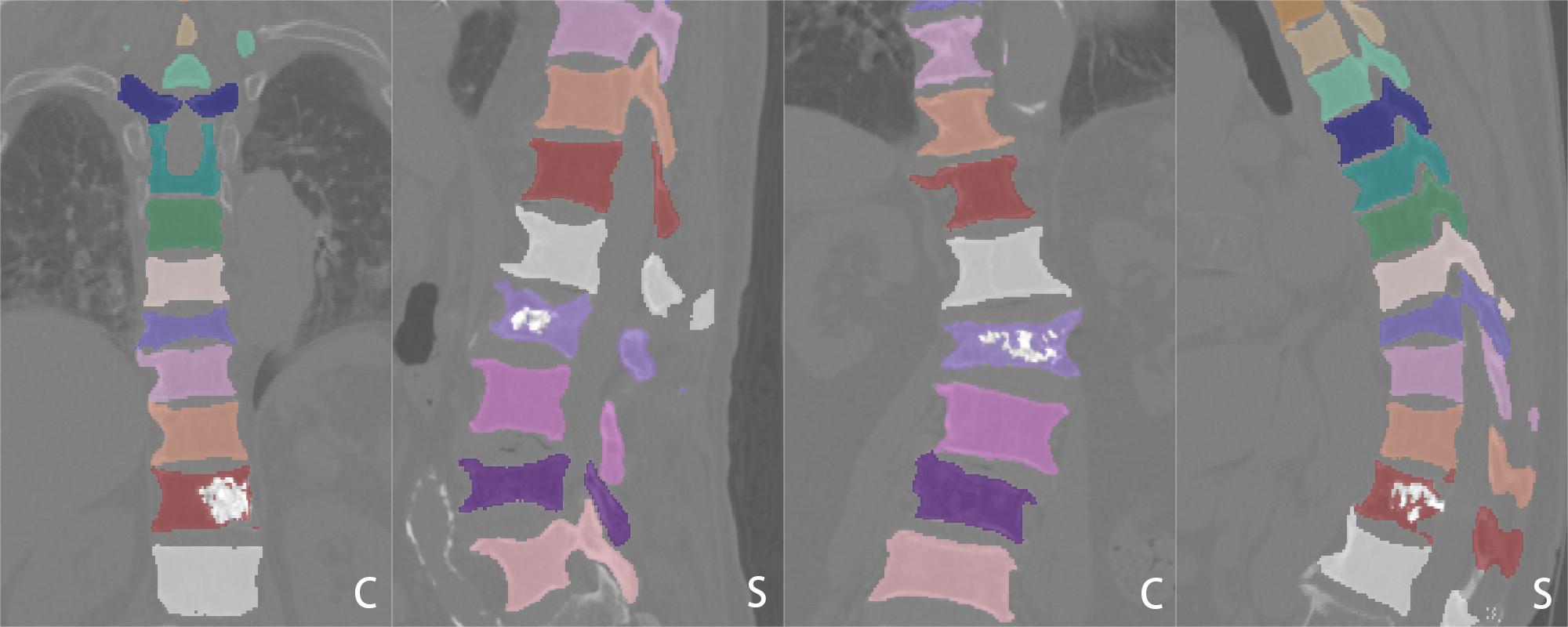}
\caption{The validation results of two images affected by artifacts, "C" and "S" denote the coronal and sagittal planes of the pictures, respectively.}
\label{fig:artifact}
\end{figure*}

We conducted the generalization experiment using model weights trained on the VerSe19 and VerSe20 datasets. SpineCLUE has not been trained on these abnormal data to evaluate the generalization performance of our method for a variety of cases in clinical practice. Fig.\ref{fig:metal-interference} illustrates the generalization effect of SpineCLUE on the metal implant images in the dataset. The quantity of metallic steel nails is typically indeterminate, and their Hounsfield Unit (HU) values vary significantly from those of tissue and bone. Although the quality of the segmentation mask in the lumbar spine images is suboptimal, our approach shows the ability to stably localize and identify the vertebrae. Fig.\ref{fig:bend-whole} depicts the results of identifying the four cases of scoliosis. Each case exhibits a different degree of scoliotic lesion, which distinguishes the vertebral segment from the normal segment in terms of spatial positioning. Some spines also display artifacts on specific vertebrae as shown in Fig.\ref{fig:artifact}, which can impede the identification of vertebrae. In contrast to metal implants, the artifacts are typically unevenly distributed and appear sporadically within a local area of the vertebrae. Without requiring training on anomalous data, SpineCLUE successfully generalizes the identification results for the aforementioned cases, which are common in clinical practice and not well addressed by current methods. The results of our method on the abnormal dataset demonstrated an ID-rate of 96.28\%.

\subsection{Visualization with t-SNE}
\begin{figure}[h]
    \centering
    \subfigure[3D-ResNet]{
        \includegraphics[width=0.22\textwidth]{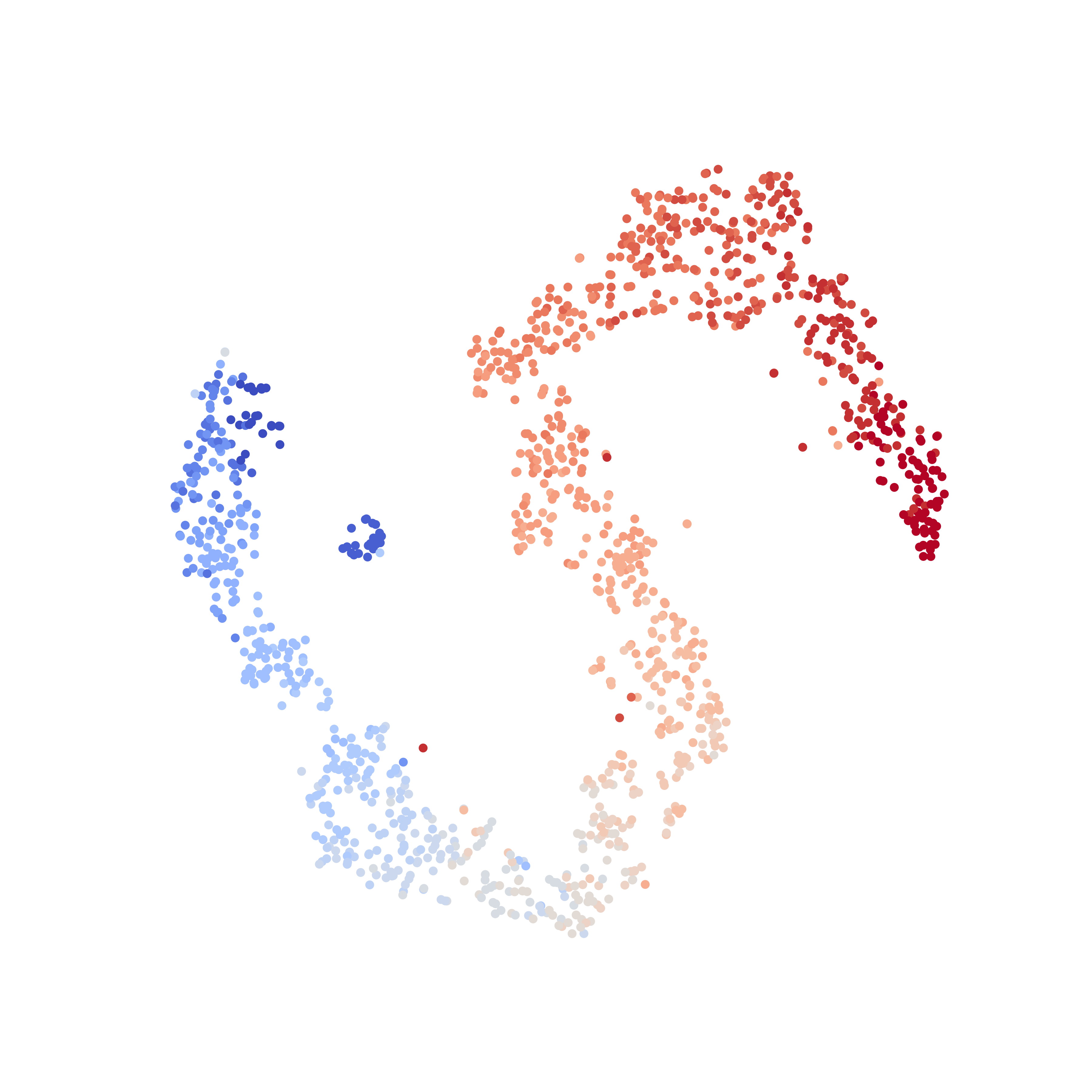}
        \label{fig:3D-ResNet}
    }
    \subfigure[3D-ResNeXt]{
        \includegraphics[width=0.22\textwidth]{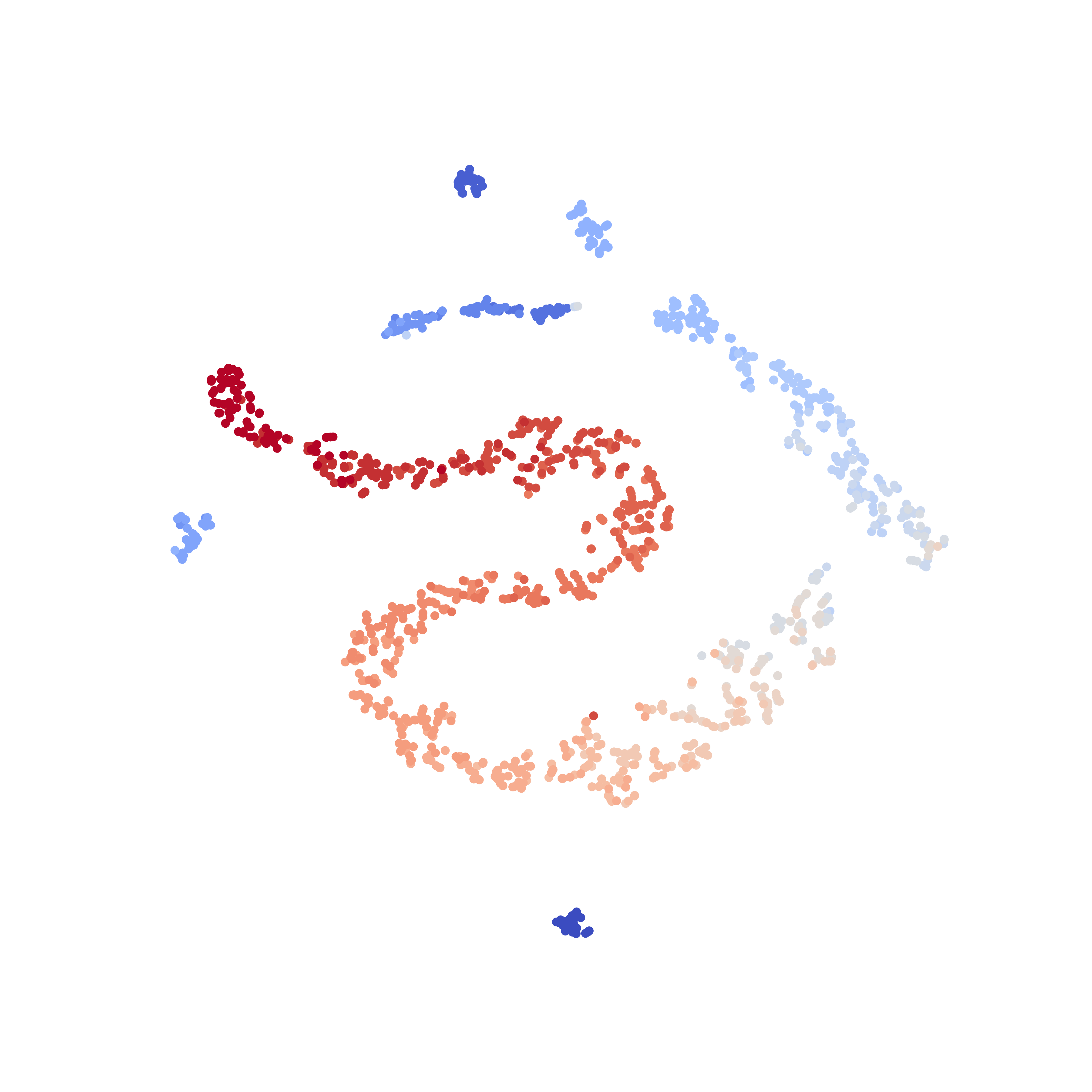}
        \label{fig:3D-ResNeXt}
    }
    \subfigure[3D-ViT]{
        \includegraphics[width=0.22\textwidth]{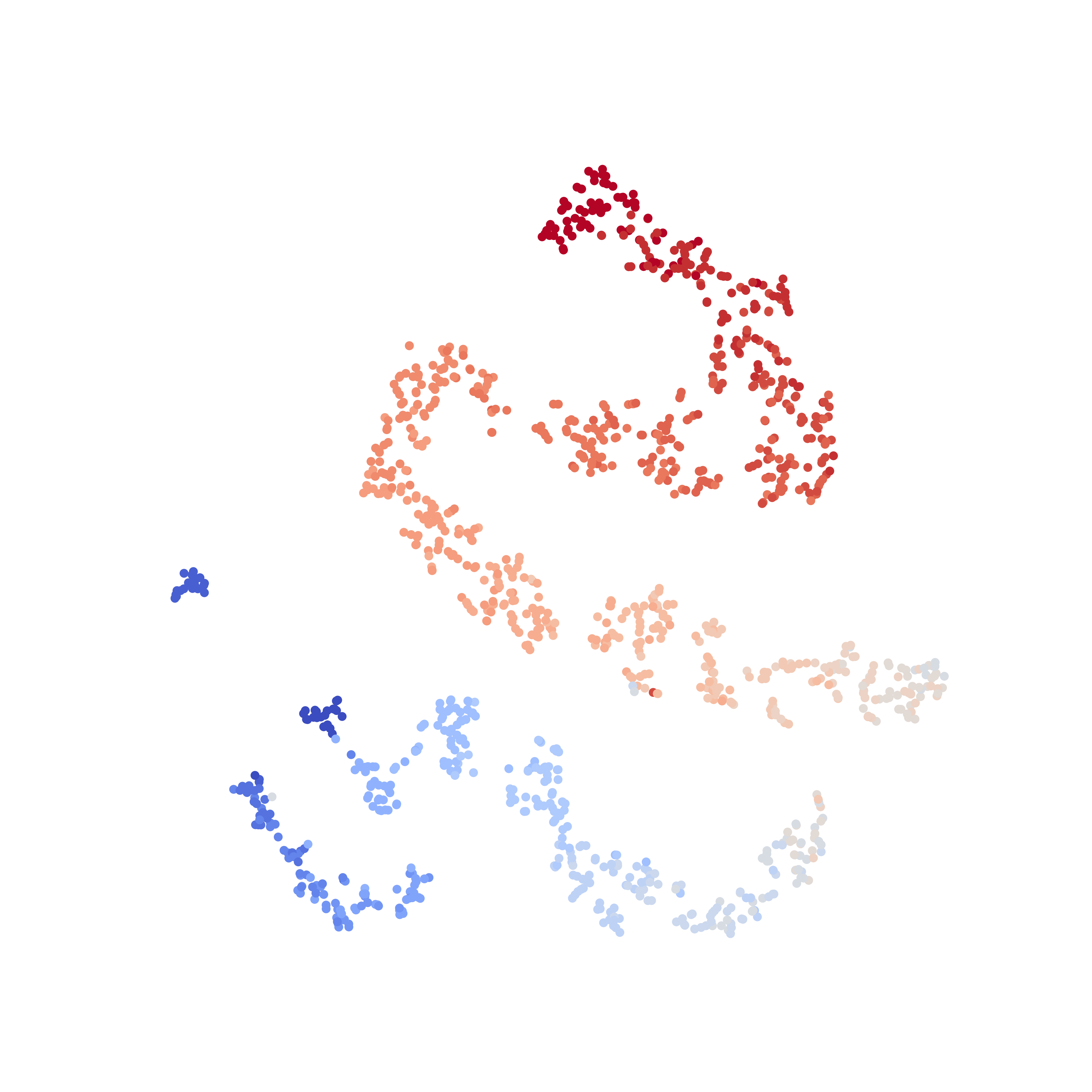}
        \label{fig:ViT}
    }
    \subfigure[SpineCLUE]{
        \includegraphics[width=0.22\textwidth]{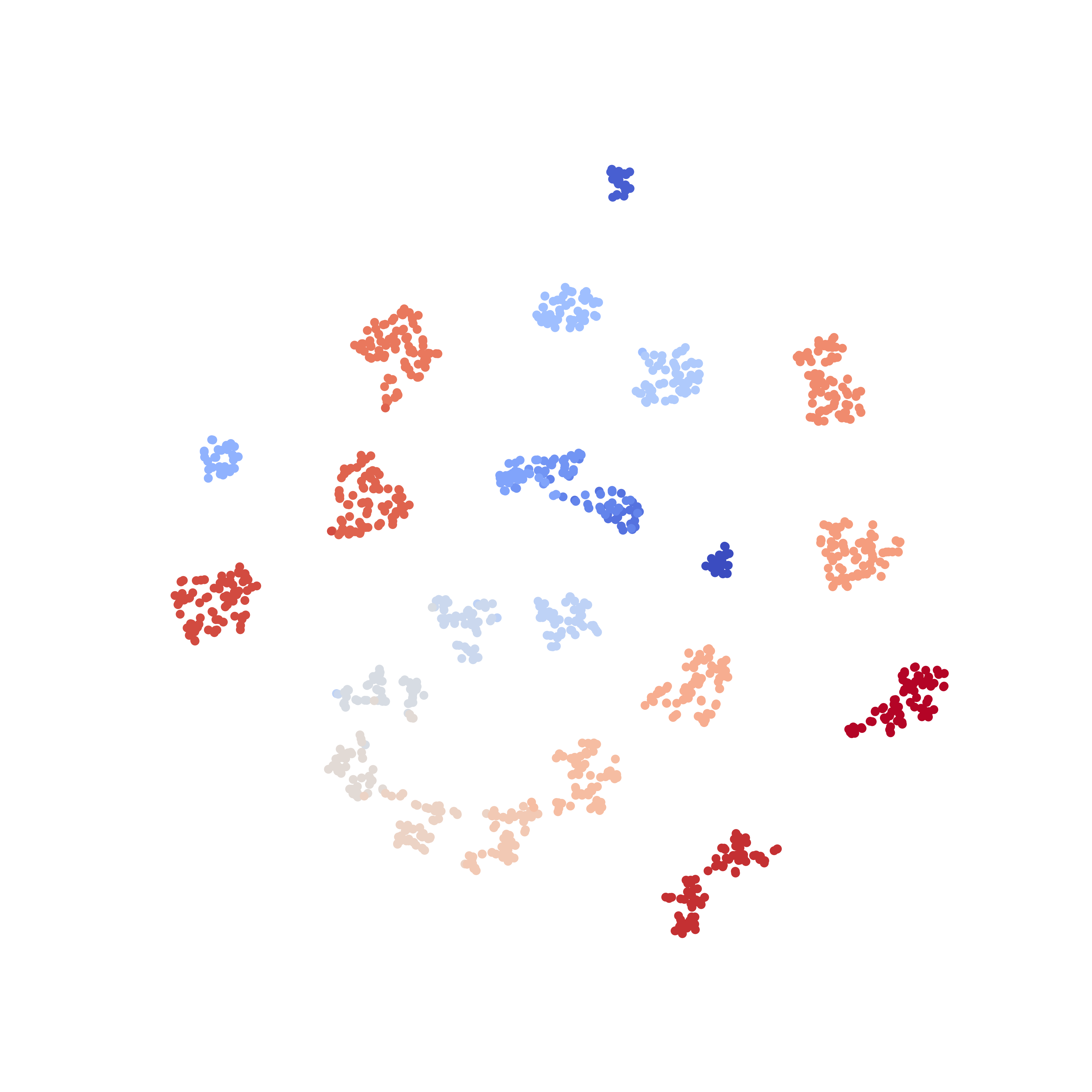}
        \label{fig:SpineCLUE}
    }
    \caption{The visualizations of t-SNE on VerSe20. Different colors are used to distinguish the vertebrae. The closer spatial distance between color blocks indicates a poorer representation of the model.}
    \label{fig:clustering}
\end{figure}

We use t-SNE \cite{t-sne} to visualize the clustered feature representations of the vertebrae identification network from 3D-ResNet-101 \cite{3DResNet}, 3D-ResNext \cite{xie2017aggregated}, 3D-ViT \cite{dosovitskiy2020image}, SpineCLUE respectively in Fig.~\ref{fig:clustering}. We clustered the embeddings of the backbone, without performing the projection head part of each model. SpineCLUE demonstrates higher distinctiveness among neighboring vertebrae, with minimal adhesion. Whereas the other networks are less able to alienating the neighboring vertebrae in feature space.

\section{Discussion}
The key differentiation between SpineCLUE and the previous research is the removal of assumptions about the number of vertebrae in the image. This is achieved through the implementation of a three-phase recursive framework at the vertebrae-level instead spine-level. To solve the restricted amount of information contained in the box, We propose a novel vertebrae-level paradigm for vertebrae identification that performs localization, segmentation, and identification sequentially instead of independently.
It is noteworthy that Liao et al.\cite{Liao2018} also addresses this issue by conducting a joint search for the labels of all vertebrae within a limited search space. This enables the exchange of global messages among the vertebrae. Chen et al.\cite{chen2020hmm} incorporates a Hidden Markov Model to impose explicit spatial and sequential constraints, thereby enhancing robustness and interpretability. Both Liao et al.\cite{Liao2018} and Chen et al.\cite{chen2020hmm} concentrate on extracting spinal morphology while fine-tuning the recognition sequence for aggregating global messages. Instead of emphasizing on the overall spinal morphology, SpinCLUE implicitly aggregates global information by means of message aggregation between vertebrae. 
In contrast to the approach taken by Liao et al.\cite{Liao2018}, where the vertebrae enclosing box was directly inputted into the bidirectional LSTM to capture long-distance contextual information, our method incorporates the uncertainty score of each vertebra. This additional information aims to improve the model's capacity to effectively perceive long-distance information. 

We have also analyzed the running time of the framework. For cervical and lumbar CT data with fewer vertebrae bodies, the average running time of our method is 22 seconds, while for full spine or large area chest data, the average running time is 40 seconds. This relies heavily on our framework, which eliminates the need for iterative optimization in order to predict the vertebral classes, high-quality mask, and localization information for the vertebrae in a single step. In contrast to vertebrae identification at vertebrae-level which typically takes several minutes, our method has a clear advantage in speed. However, in comparison to methods that can meet real-time demands at the spine-level, our method still has the potential to further decrease the running time. According to experimental statistics, the re-sampling of 3D CT data, cold start loading of the YOLO model, and post-processing and saving of prediction results have been identified as the most time-consuming tasks in SpineCLUE. It is possible to further reduce the model parameter size of the backbone network used for localization, segmentation, and identification, which will also be a future research direction. The message fusion module is capable of operating without incurring additional time overhead, thanks to its fewer parameters and computational load.

We conducted a validation of SpineCLUE on two public datasets, VerSe19 and VerSe20. Additionally, we compared the performance of SpineCLUE with existing state-of-the-art methods. For each dataset, all models employ identical dataset partitioning and evaluation metrics. SpineCLUE demonstrates superior ID-rate performance on both datasets, achieving scores of 97.92\%, 98.14\%, 98.73\% and 97.87\% on the public test set and the hidden test set, respectively. In contrast to other models for evaluating the performance of abnormal vertebrae body detection in the VerSe dataset through cross-validation, we employed an abnormal dataset as a separate test set for validation purposes. The dataset comprises of anomalous vertebrae bodies with arbitrary FOVs, including metal implants, spinal scoliosis, and vertebrae compression fractures. The results of our method on the abnormal dataset demonstrated an ID-rate of 96.28\%, suggesting its generalization ability on abnormal cases.

Despite achieving high recognition rate scores, SpineCLUE exhibits certain flaws and shortcomings. As shown in Table \ref{tab:different-vertebrae}, when comparing the accuracy of identification between the lumbar and thoracic spine, which have severe inter-class similarity and intra-class variability, it is evident that there is room for improvement in the identification of the cervical spine. Cervical vertebrae pose challenges in accurately encapsulating them using a bounding box, primarily due to their variations in size and shape. Future work will include learning features that are more robust to cervical vertebrae.

\section{Conclusion}
In this paper, we propose a three-stage end-to-end method to address the challenges and limitations inherent in vertebrae identification. We introduce dual-factor density clustering algorithm to obtain localized information for individual vertebra, facilitating the subsequent identification process. Specifically, a vertebrae contrastive learning strategy is futher introduced to alleviate challenges related to inter-class similarity and intra-class variability. Additionally, we employ uncertainty message fusion
to refine the identification results by leveraging the sequential features. The experimental results on public datasets substantiate the superior performance of SpineCLUE. We anticipate that SpineCLUE will play a crucial role in the field of medical image analysis, offering heightened accuracy and reliability for clinical applications.

\bibliography{mybib}

\end{document}